\documentclass[10pt,twocolumn,letterpaper]{article}

\usepackage{cvpr}      %

\usepackage{graphicx}
\usepackage{amsmath}
\usepackage{mathtools}
\usepackage{amssymb}
\usepackage{booktabs}
\usepackage{multirow}

\usepackage[font=small,skip=4pt]{caption}
\usepackage{xspace}
\usepackage[english]{babel}
\usepackage{array}
\usepackage[accsupp]{axessibility}  %

\newcolumntype{P}[1]{>{\centering\arraybackslash}p{#1}}
\newcolumntype{C}[1]{>{\centering\arraybackslash}m{#1}}
\newcolumntype{L}[1]{>{\centering\raggedright\arraybackslash}m{#1}}

\makeatletter
\DeclareRobustCommand\onedot{\futurelet\@let@token\@onedot}
\def\@onedot{\ifx\@let@token.\else.\null\fi\xspace}
\def\eg{\emph{e.g}\onedot}

\def\etal{\emph{et al}\onedot}
\DeclareMathAlphabet{\mathcal}{OMS}{cmsy}{m}{n}
\newcommand{\ve}[1]{\ensuremath{\mathbf{#1}}} %
\newcommand{\set}[1]{\ensuremath{\mathcal{#1}}} %
\usepackage[dvipsnames]{xcolor}

\usepackage[pagebackref,breaklinks,colorlinks]{hyperref}

\usepackage[capitalize]{cleveref}
\crefname{section}{Sec.}{Secs.}
\Crefname{section}{Section}{Sections}
\Crefname{table}{Table}{Tables}
\crefname{table}{Tab.}{Tabs.}

\def\cvprPaperID{6939} %
\def\confName{CVPR}
\def\confYear{2022}

\begin{document}

\title{Feature-Style Encoder for Style-Based GAN Inversion}

\author{Xu Yao$^{1,2}$, Alasdair Newson$^{1}$, Yann Gousseau$^{1}$, Pierre Hellier$^{2}$ \\
$^{1}$ LTCI, T\'el\'ecom Paris, Institut Polytechnique de Paris, France\\
$^{2}$ InterDigital R\&I, 975 avenue des Champs Blancs, Cesson-S\'evign\'e, France\\
}

\twocolumn[{%
\renewcommand\twocolumn[1][]{#1}%
\maketitle
\small
\setlength{\tabcolsep}{0pt}
\renewcommand{\arraystretch}{0.5}
\centering
\begin{tabular}{P{0.124\linewidth}P{0.124\linewidth}P{0.124\linewidth}P{0.124\linewidth}P{0.124\linewidth}P{0.124\linewidth}P{0.124\linewidth}P{0.124\linewidth}}
\centering
Source & 
Optimization \cite{abdal2020image2stylegan++} & In-domain \cite{zhu2020domain}& pSp\cite{richardson2020encoding} & e4e\cite{tov2021designing} & restyle-pSp \cite{alaluf2021restyle} & HFGI\cite{wang2021high} & Ours
\\
\multicolumn{8}{c}{\includegraphics[width=0.99\linewidth]{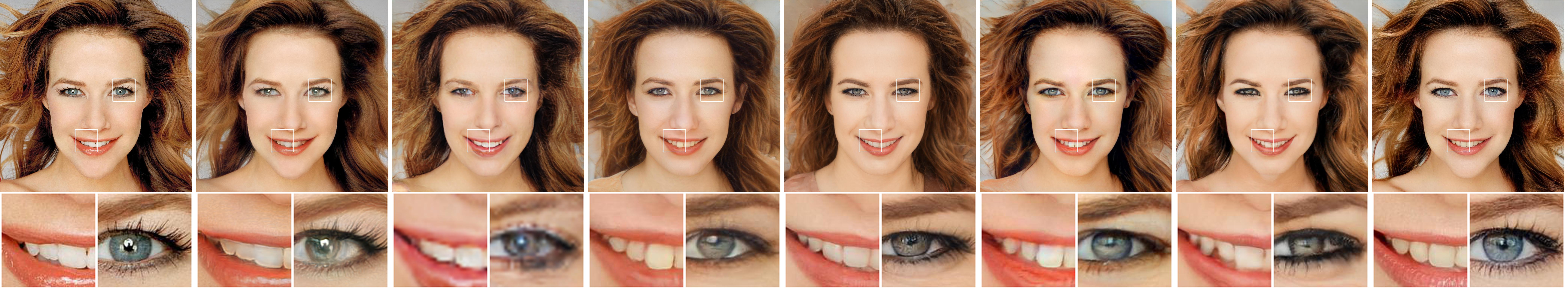}}
\\
LPIPS $\downarrow$ & 0.080 & 0.134 & 0.143 & 0.177 & 0.125 & 0.109 & \textbf{0.064}
\\
\end{tabular}
\label{teaser}
\captionof{figure}{\textbf{Inversion of a real image in the latent space of StyleGAN2.} 
We compare our model against state-of-the-art for the inversion of StyleGAN2 \cite{karras2019analyzing} pre-trained on face domain. 
Our method outperforms the state-of-the-art by up to $20\%-50\%$ in LPIPS distance\cite{zhang2018unreasonable}.\\}%
}]

\begin{abstract}
We propose a novel architecture for GAN inversion, which we call Feature-Style encoder. The style encoder is key for the manipulation of the obtained latent codes, while the feature encoder is crucial for optimal image reconstruction. 
Our model achieves accurate inversion of real images from the latent space of a pre-trained style-based GAN model, obtaining better perceptual quality and lower reconstruction error than existing methods. Thanks to its encoder structure, the model allows fast and accurate image editing. 
Additionally, we demonstrate that the proposed encoder is especially well-suited for inversion and editing on videos.
We conduct extensive experiments for several style-based generators pre-trained on different data domains.
Our proposed method yields state-of-the-art results for style-based GAN inversion, significantly outperforming competing approaches. Source codes are available at \url{https://github.com/InterDigitalInc/FeatureStyleEncoder}.
\end{abstract}

\section{Introduction}
\label{sec:intro}

The incredible image synthesis power of Generative Adversarial Networks (GANs)\cite{Goodfellow2014} has been amply demonstrated by the great quantity of work on this architecture since its creation. However, since a GAN only decodes an image from a probabilistic latent space, a significant research problem is how to \emph{encode} images into the latent space of a pretrained GAN, especially in the case of real (photographic) images, as opposed to \emph{synthetic} images, which are generated by sampling in the latent space. Recent studies \cite{shen2019interpreting,harkonen2020ganspace,wu2020stylespace,shen2021closed} have shown that it is possible to control semantic attributes of synthetic images by manipulating the latent space of a pre-trained GAN, however an efficient encoding method, necessary for real images, still remains an open problem, especially in the case of these editing tasks.
Thus, an ideal encoder should lead to high perceptual quality on real images, while ensuring that the inverted latent codes are amenable to the same editing that is possible in the case of synthetic images.

Among the many studies on GAN inversion, recent works have been primarily focused on style-based generators \cite{karras2019style, karras2019analyzing, karras2021alias}, because of their excellent performance in high quality image synthesis, especially on faces. Unlike traditional generative models which feed the latent code though the input layer only, a style-based generator feeds latent code through adaptive instance normalization at each convolution layer to control the style of the generated image.

To project a given image to the latent space of a style-based GAN model, there are two approaches: optimization and learning an encoder. The most straight-forward is to perform optimization \cite{abdal2020image2stylegan++} on the latent code by minimizing the image reconstruction error. In order to achieve higher perceptual quality, including the feature maps in the optimization has been proposed \cite{zhu2021barbershop,kang2021gan}.
In spite of these achievements, optimization-based methods have significant shortcomings. One major drawback is that the optimization process is carried out locally with respect to a single image. Thus, the resulting inverted latent codes do not necessarily lie on the original latent space, which makes them difficult to use for editing tasks, as shown by \cite{tov2021designing,alaluf2021restyle}.

A better approach is to \emph{learn an encoder} to invert images to latent codes \cite{zhu2020domain, richardson2020encoding, tov2021designing, alaluf2021restyle, wei2021simple}. The inverted latent codes are more regularized and therefore \emph{better suited for editing}. The inversion process is also much faster, which is especially important for computationally intensive tasks such as video editing. However, current encoder-based methods also have several limitations. Firstly, the reconstruction error of the inversion is larger than that of optimization-based methods. Secondly, the reconstructed image is perceptually similar to the input but lacks finer details and appears over smoothed. %
Thirdly, encoding only the latent code fails to generate correct inversion on outlier data. For instance, given a talking face video, we observe that such methods fail to invert non-frontal poses, thus damaging the consistent reconstruction along the sequence. %

To tackle the above mentioned weaknesses, we propose an new inversion architecture. We learn an encoder in the \emph{feature-style} space, which maps an image to a feature code and a latent code. The feature code encodes spatial details, and the latent code is used for editing. %
This design significantly improves the perceptual quality of the inversion and achieves a balanced trade-off between reconstruction quality and editing capacity. 
The main contributions of our paper can be summarized as follows:
\begin{itemize}
\setlength\itemsep{-0.2em}
    \item We propose a new GAN encoder, which is the first to exploit the idea of encoding feature and style separately, without any optimization step at inference time. Such optimization is both costly and leads to unreliable editing results. %
    Our Feature-Style encoder, on the other hand, significantly improves the perceptual quality of the inversion and improves latent space editing;
    \item We present a novel training approach, which learns two inversions simultaneously - one which is amenable to editing and one of higher fidelity but less adapted for editing. By training in this manner, our encoder achieves a balanced trade-off between reconstruction quality and editing capacity;
    \item We conduct extensive experiments to show that our model greatly outperforms state-of-the-art methods on inversion and editing tasks on images and videos. In particular, we improve the perceptual metrics by a very large margin ($50\%$). 
    In addition, we show that the video inversion results of our method is more consistent and stable, which favors further editing on videos.
\end{itemize}

\section{Related works}
\label{sec:related}

The goal of our work is to learn an encoder for projecting real images to the latent space of a pre-trained style-based generator. Much of the recent literature on GAN inversion pays particular attention to style-based generators \cite{karras2019style, karras2019analyzing,karras2020ada,karras2021alias}, as their latent spaces are better disentangled and have improved editing properties. 

\noindent \textbf{Style-based Generator.} \quad Karras \etal proposed the first style-based generator, named StyleGAN \cite{karras2019style}. Unlike traditional generative models which feed the latent code though the input layer only, a style-based generator feeds latent code through adaptive instance normalization at each convolution layer to control the style of the generated image. 
The perceptual quality and variety of the StyleGAN synthetic images surpassed previous image generative models \cite{karras2018progressive, biggan}. 
In StyleGAN2 \cite{karras2019analyzing}, the image quality was improved further by introducing
weight demodulation and path length regularization and redesigning the generator normalization. The StyleGAN2-Ada \cite{karras2020ada} explored the possibility to train a GAN model with limited data regimes, by using an adaptive discriminator augmentation mechanism that significantly stabilizes training. The third generation, alias-free GAN \cite{karras2021alias}, addressed the aliasing artifacts in the generator, by employing small architectural changes to discard unwanted information and boost the generator to be fully equivariant to translation and rotation.

\noindent \textbf{Latent Space Editing.} \quad The motivation of GAN inversion is to achieve real image editing using the latent space of a pretrained GAN model. Various studies show it possible to edit synthetic images by manipulating the corresponding latent code. Local semantic editing can be achieved by optimizing the latent code directly \cite{abdal2020image2stylegan++,ling2021editgan}. To explore high level semantic information in the latent space, learning based techniques have been proposed. These techniques include unsupervised exploration\cite{voynov2020unsupervised}, learning linear SVM models\cite{shen2019interpreting}, principle component analysis on the latent codes\cite{harkonen2020ganspace}, and k-means clustering of the activation features\cite{collins2020editing}. To achieve better disentangled editing, \cite{abdal2020styleflow,patashnik2021styleclip,yao2021latent,tewari2020stylerig,hou2021guidedstyle} proposed to learn neural networks in the latent space. The recent works \cite{shen2021closed,wang2021geometry} discovered interpretable transforms by directly decomposing the weights or feature maps of pre-trained GANs. Additionally, \cite{alharbi2020disentangled,kwon2021diagonal} modify the style-based GAN architecture and retrain it for better disentanglement in image generation.  \cite{pidhorskyi2020adversarial,park2020swapping,kim2021exploiting} train jointly an encoder and a style-based decoder architecture for image manipulations. 

\begin{figure*}[t]
\centering
\includegraphics[width=0.9\linewidth]{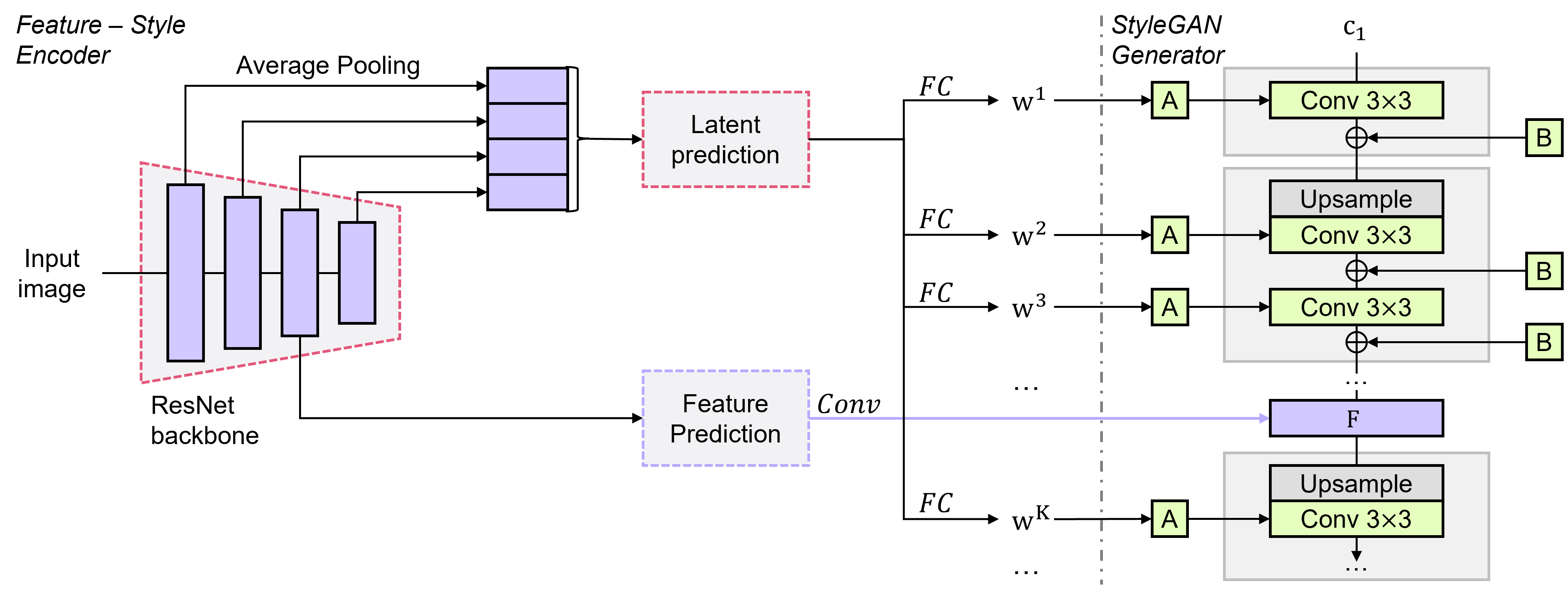}
\caption{\textbf{Feature - Style Inversion Architecture.} Our model consists of a ResNet backbone and two output branches: one for latent code prediction, the other for feature code prediction. The inverted latent code \ve{w} is a concatenation of $N$ latent blocks $\{\ve{w}^1,\ve{w}^2, ..., \ve{w}^N\}$, each controlling a separate convolution layer in the generator. During generation, we replace the feature maps at the $K^{th}$ convolution layer of the generator with the inverted feature code $\ve{F}$, and synthesize the inversion with the latent blocks $\{\ve{w}^K, ..., \ve{w}^N\}$. $K$ is a fixed parameter, chosen so that reconstruction is accurate and editing can be performed efficiently.
}
\label{arch}
\end{figure*}

\noindent \textbf{GAN Inversion.} \quad 
The goal of GAN inversion is to encode a real image to the latent space of a pretrained generator, so that the image generated from the inverted latent code is the reconstruction of the input image. Among the rich literature on GAN inversion \cite{xia2021gan}, the approaches addressing style-based generators can be classified into three types: optimization based methods \cite{abdal2019image2stylegan,huh2020transforming,xu2021continuity,zhu2021barbershop,kang2021gan}, encoder based models \cite{richardson2020encoding,tov2021designing,alaluf2021restyle,wang2021high,wei2021simple} and hybrid methods \cite{zhu2020domain,chai2021ensembling,yu2021adaptable}. The optimization based methods update directly the inverted latent code by minimizing the reconstruction error on the input image. For StyleGAN inversion, Abdal \etal \cite{abdal2019image2stylegan} proposed to embed the input image in an extended latent space \set{W^+}, which offers better flexibility and improves the reconstruction quality. Recently, it was shown that including a feature code in the optimization helps preserve more spatial details and improves the perceptual quality of inversion \cite{zhu2021barbershop,kang2021gan}.
Despite the satisfying reconstruction quality, optimization-based methods usually present lower editing quality due to the randomness in the optimization process. To better regularize the inversion, encoder based methods train an encoder to map real images to the latent space of the pretrained generator. Richardson \etal \cite{richardson2020encoding} proposed the first baseline to learn an encoder for StyleGAN inversion. To improve editing capacity, Tov \etal \cite{tov2021designing} proposed a regularization term which forces the inverted latent code in \set{W^+} to lie closer to the original latent space. A recent concurrent work of Wang \etal \cite{wang2021high} formulated the inversion task to a data compression problem and proposed an adaptive distortion alignment module to improve the reconstruction quality. Encoder-based methods achieve better editing results but degrade reconstruction quality. On the other hand, hybrid methods take the inverted latent code from a pretrained encoder as initialization and perform optimization on it. Zhu \etal \cite{zhu2020domain} proposed to learn a domain-guided encoder and use it as a regularizer for domain-regularized optimization. However, despite the gain in the reconstruction quality, the optimization step makes hybrid methods less suited for video inversion and editing.

\section{Method}
\label{sec:method}

In this section, we introduce the Feature-Style encoder for real image inversion and editing via the latent space of a pretrained style-based generator. In our model, we use a ResNet backbone with two output branches: one for the inverted latent code, the other for the encoded feature maps. Figure \ref{arch} shows the overall architecture, as well as how the latent code and feature maps are plugged into the StyleGAN architecture to generate the reconstructed image.

\subsection{Overview}

A style-based generator, such as StyleGAN \cite{karras2019style,karras2019analyzing,karras2021alias}, consists of a mapping network and a generator \ve{G}. The mapping network first maps a random latent code $\ve{z} \in \set{Z}$ to an intermediate latent code $\ve{w} \in \set{W}$, which is further used to scale and bias the feature maps, ie the outputs of a convolutional layer, after each layer in the generator. To project a synthetic image $\ve{G(w)}$ to the latent space, it is possible to compute the latent code in the original latent space \set{W} and achieve a satisfying inversion. However, it is much more difficult to project a real image to the original latent space \cite{karras2019analyzing}, due to the gap between the real data distribution and the synthetic one. %
An alternative is to project real images to an extended latent space \set{W^+} \cite{abdal2019image2stylegan}, where $\ve{w} \in \set{W^+}$ is a concatenation of $N$ latent blocks $\{\ve{w}^1, \ve{w}^2, ..., \ve{w}^N \}$, each controlling a convolution layer in the generator.

In addition to the latent code, StyleGAN1-2 \cite{karras2019style,karras2019analyzing} add Gaussian noise after each convolution layer to generate local spatial variations. Abdal \etal \cite{abdal2020image2stylegan++} show that it is possible to reconstruct a real image by jointly optimizing the latent code in \set{W^+} and the noise maps. However, in the original StyleGAN architecture, these noise maps are drawn from a Gaussian distribution, and are therefore not designed to represent geometric elements of the image. Unfortunately, this optimization of noise maps ends up encoding such geometric information in the noise. This is particularly problematic for latent space editing, since it becomes very difficult to modify geometric information efficiently, which is extremely limiting for an editing algorithm.

On the other hand, without optimization on the noise maps, it is impossible to achieve perfect reconstruction for real images by optimizing the latent code only. To tackle this dilemma, the authors of \cite{zhu2021barbershop,kang2021gan} propose to include the \emph{feature maps} in the optimization process, rather than optimizing noise maps, in order to preserve spatial details. Performing optimization on both the latent code and feature maps yields near perfect reconstruction on real images.

In our work, we aim to have the best of both worlds: we wish to achieve this high reconstruction fidelity, while maintaining the speed and editing capacity of an encoder. Thus, we propose our \emph{Feature-Style encoder}, which projects an image to a latent code $\ve{w} \in \set{W^+}$, and
a \emph{feature code} $\ve{F} \in \set{F} \subset \mathbb{R}^{h\times w \times c}$. This feature code is thus a tensor, and replaces the original GAN's feature map at a fixed layer indexed $K$ of the generator. The parameters $(h,w,c)$ correspond to the spatial size and the number of channels of the tensor, and depend on the layer $K$. The rest of the layers are controlled by the latent code $\ve{w}$. We now present this architecture in more detail.

\subsection{Feature-Style encoder}

 \noindent \textbf{Encoding} \quad The basic structure of our Feature-Style encoder is modelled on the classic approach used by most previous works on style-based GAN inversion \cite{richardson2020encoding,tov2021designing,alaluf2021restyle,wei2021simple,wang2021high}, which employ a ResNet backbone. After this, as shown in Figure \ref{arch}, we have two output branches: a latent prediction branch to encode the latent code $\ve{w} \in \set{W^+}$, and a feature prediction branch to encode the feature code $\ve{F} \in \set{F}$. The ResNet backbone contains four blocks, each down-sampling the input feature maps by a factor of $2$. Given an input image, we extract the feature maps after each block. In the latent branch, the four groups of feature maps are passed through an average pooling layer, concatenated and flattened to produce the latent prediction. This is then mapped to the latent code $\ve{w} = \{\ve{w}^1, \ve{w}^2, ..., \ve{w}^N \}$. Each latent block $\ve{w}^i$ is generated from a different mapping network, expressed by a single fully connected layer. In the feature branch, the feature maps extracted after the penultimate block are passed through a convolutional network to encode the feature code \ve{F} (see Figure~\ref{arch}). 
Let $\ve{G}^K(\ve{w})$ denote the feature maps at the $K^{th}$ convolution layer of the generator. To generate the inversion, we replace $\ve{G}^K(\ve{w})$ with the feature code \ve{F}, and use the rest of the latent codes $\{\ve{w}^K, ..., \ve{w}^N \}$ to generate \ve{G(w, \ve{F})}. We choose $K = 5$ for a balanced trade-off between the inversion quality and editing capacity, leading to $\set{F} \subset \mathbb{R}^{16 \times 16 \times 512}$.

\noindent \textbf{Editing} \quad 
In a style-based generator, the styles corresponding to coarse layers control high-level semantic attributes, the styles of the middle layers control smaller scale features, whereas the finer styles control micro structures. 
Given an input latent code $\ve{w}$, let us consider that we have a latent code $\tilde{\ve{w}}$ corresponding to a desired editing, where $\tilde{\ve{w}}$ is obtained from a latent space editing method \cite{shen2019interpreting,harkonen2020ganspace,shen2021closed}. %
To include the edits controlled by $\{\ve{w}^1, ..., \ve{w}^{K-1} \}$, it is necessary to modify the feature code. 
Thus we generate two images, $\ve{G}(\ve{w})$ and $\ve{G}(\tilde{\ve{w}})$, from $\ve{w}$ and $\tilde{\ve{w}}$, respectively. During generation, we extract the input features at the $K^{th}$ convolution layer $\ve{G}^K(\ve{w})$ and $\ve{G}^K(\tilde{\ve{w}})$, compute the difference between them and add it to the encoded features \ve{F}.
The modified feature $\tilde{\ve{F}}$ is determined as:
\begin{equation}
\label{eq:modif_feature}
    \tilde{\ve{F}} = \ve{F} + \ve{G}^K(\tilde{\ve{w}}) - \ve{G}^K(\ve{w}).
\end{equation}
Then we generate the edited image $\ve{G}(\tilde{\ve{w}},\tilde{\ve{F}})$ with $\tilde{\ve{w}}$ and the modified feature code $\tilde{\ve{F}}$.

\section{Training}
\label{sec:training}

\begin{figure}[t]
\centering
\includegraphics[width=0.99\linewidth]{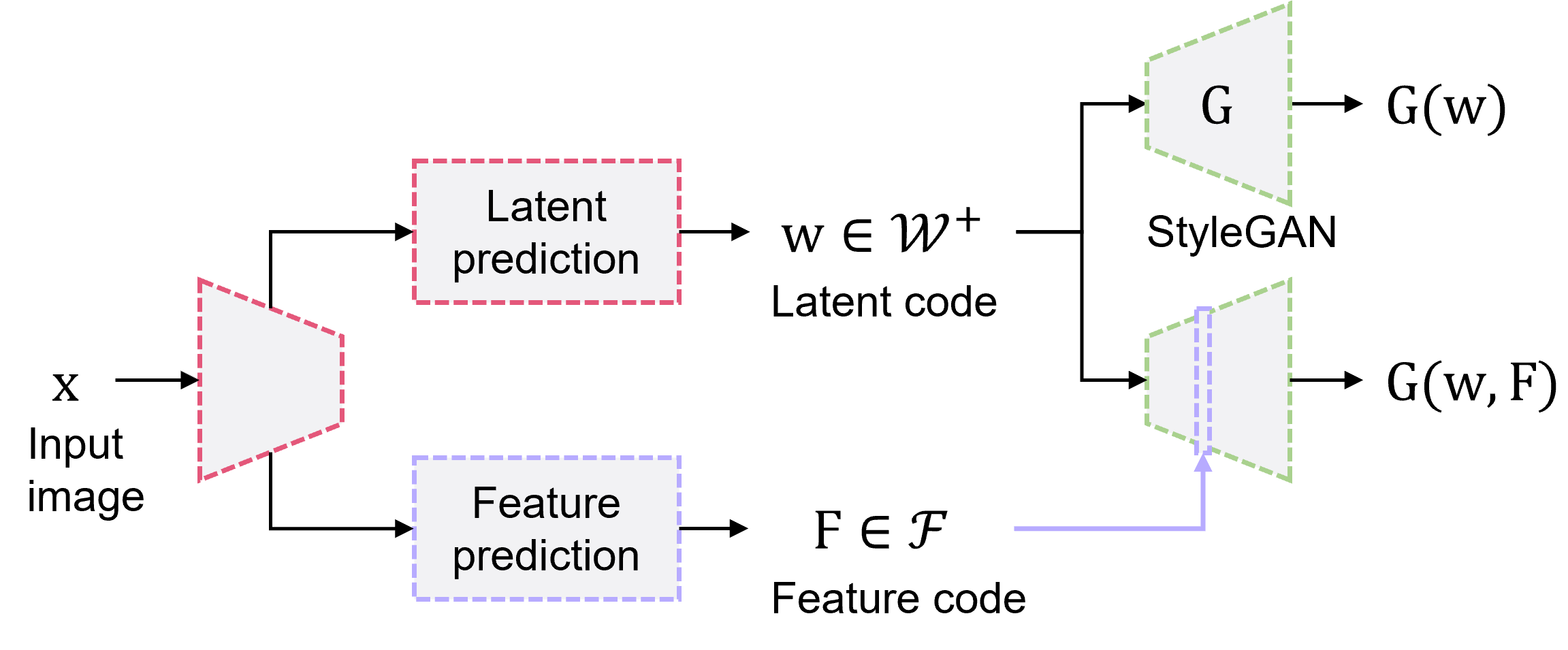}
\caption{\textbf{Training approach.} During training, the encoder learns two inversions simultaneously - one generated with only the latent code, the other generated with both outputs. This design ensures the editing capacity of the learned latent code. %
}
\label{arch_training}
\end{figure}

\subsection{Training data} 

Previous methods on GAN inversion \cite{richardson2020encoding,tov2021designing,alaluf2021restyle,wang2021high,wei2021simple} take only real image datasets as training data. However, compared with the synthetic images, the perceptual quality of the images resulting from the inversion is worse. An intuitive explanation is that there is a difference between the data distributions of real and synthetic images. The encoder is trained to project a given image to the extended latent space \set{W^+}. If synthetic images are not viewed by the encoder, the resulting latent code may not perform as well as those of the original latent space. Therefore, we include both synthetic and real images as training data. 

In the case of StyleGAN2 \cite{karras2019style,karras2019analyzing}, the generator accepts two inputs, a latent code and a group of noise maps for local variations. A synthetic image is generated from a random latent code and a group of random noises. During training, if the input is a synthetic image, we pass the ground truth noises into the generator, so that the encoder can focus on the information encoded by the latent code. If the input is a real image, we pass random noises into the generator. Note that in this case even if the latent code is perfectly inverted, local variations exist between the inversion and the input.

\setlength{\abovedisplayskip}{4pt}

\subsection{Losses} 

As shown in Figure \ref{arch_training}, our Feature-Style encoder inverts an input image \ve{x} to a latent code $\ve{w}$, and a feature code $\ve{F}$. To ensure the full editing capacity of the latent code, the encoder is trained on two inversions simultaneously - one generated with only the latent code $\tilde{\ve{x}}_1=\ve{G(w)}$ and the other generated with both the feature code and the latent code $\tilde{\ve{x}}_2=\ve{G(w, \ve{F})}$.

\noindent \textbf{Pixel-wise reconstruction loss} \quad In the case of a synthetic image, the reconstruction is measured using mean squared error (MSE) on $\tilde{\ve{x}}_1$ only. In this special case, the ground-truth latent code exists and the feature map is irrelevant. For real image input, the ground-truth latent code is unknown, so a per-pixel constraint may be too restrictive. The loss is expressed as:
\begin{equation}
\label{eq:mse}
    \mathcal{L}_{mse} = 
     ||\ve{G(w)} - \ve{x}||_2.
\end{equation}
\noindent \textbf{Multi-scale perceptual loss (LPIPS)} \quad A common problem of the previous GAN inversion methods is the lack of sharpness of the reconstructed image, despite using the per-pixel MSE. To tackle this, we propose a multi-scale loss design which enables the reconstruction of finer details. Given an input image \ve{x} and its inversion $\tilde{\ve{x}}$, a multi-scale LPIPS loss is defined as:
\begin{equation}
\label{eq:m_lpips}
    \mathcal{L}_{m\_lpips}(\tilde{\ve{x}}) = 
     \sum_{\substack{i=0}}^2||\ve{V}(\lfloor\tilde{\ve{x}}\rfloor_i) - \ve{V}(\lfloor\ve{x}\rfloor_i)||_2,
\end{equation}
where $\lfloor . \rfloor_i$ refers to downsampling by a scale factor $2^i$ and \ve{V} denotes the feature extractor. This design allows the encoder to capture the perceptual similarities at different scales, which favors the perceptual quality of the inversion. This loss is applied on both inversions.

\noindent \textbf{Feature reconstruction} \quad To ensure the possibility of using Eq.(\ref{eq:modif_feature}) to edit the feature code,
$\ve{F}$ should be similar to the input feature maps at the $K^{th}$ convolution layer in the generator, denoted by $\ve{G}^K(\ve{w})$. Therefore, we propose a feature reconstruction loss, which favors the encoded features to stay close to the original latent space. This term is defined as:
\begin{equation}
\label{eq:fea_recon}
    \mathcal{L}_{f\_recon} = 
     ||\ve{F} - \ve{G}^K(\ve{w})||_2.
\end{equation}
The total loss is defined as:
\begin{equation}
\label{eq:total_loss}
    \mathcal{L}_{total} =  \mathcal{L}_{mse} + \lambda_{1}\mathcal{L}_{m\_lpips} + \lambda_{2}\mathcal{L}_{f\_recon},
\end{equation}
where $\lambda_{1} = 0.2$ and $\lambda_{2}=0.01$ are weights balancing each loss.

\noindent \textbf{Face Inversion} \quad For the inversion of a styleGAN model pre-trained on a face dataset, we adopt the multi-layer identity loss and the face parsing loss introduced by \cite{wei2021simple}.
Given an input image \ve{x} and its inversion $\tilde{\ve{x}}$, the multi-layer identity loss is defined as:
\begin{equation}
\label{eq:idloss}
    \mathcal{L}_{id}(\tilde{\ve{x}}) = \sum_{\substack{i=1}}^5 (1 - \langle\ve{R}_i(\tilde{\ve{x}}), \ve{R}_i(\ve{x})\rangle),
\end{equation}
where \ve{R} is the pre-trained ArcFace network \cite{deng2019arcface}. The multi-layer face parsing loss is defined as:
\begin{equation}
\label{eq:parsingloss}
    \mathcal{L}_{parse}(\tilde{\ve{x}}) = \sum_{\substack{i=1}}^5 (1 - \langle\ve{P}_i(\tilde{\ve{x}}), \ve{P}_i(\ve{x})\rangle),
\end{equation} 
where \ve{P} is a pre-trained face parsing model \cite{face-parsing}. These two above-mentioned losses are applied on both inversions. Hence the total loss for face inversion is:
\begin{equation}
\label{eq:final_loss}
    \mathcal{L}_{face} =  \mathcal{L}_{total} + \lambda_{3}\mathcal{L}_{id} + \lambda_{4}\mathcal{L}_{parsing},
\end{equation}
where $\lambda_{3} = 0.1$ and $\lambda_{4}=0.1$ are weights balancing the identity preserving and face parsing terms.

\section{Experiments}
\label{sec:expe}

In this section, we present the experimental setup and compare our method with state-of-the-art GAN inversion methods. We conduct the evaluation from two aspects: inversion quality and editing capacity. We also show results on videos as well as ablative studies.

\begin{figure*}[t]
\centering
\small
\setlength{\tabcolsep}{0pt}
\renewcommand{\arraystretch}{0}
\begin{tabular}{P{0.16\linewidth}P{0.16\linewidth}P{0.16\linewidth}P{0.16\linewidth}P{0.16\linewidth}P{0.16\linewidth}}
\centering
Source & pSp\cite{richardson2020encoding} & e4e\cite{tov2021designing} & restyle-pSp \cite{alaluf2021restyle} & HFGI\cite{wang2021high} & Ours
\\
&&&&&
\\
\multicolumn{6}{c}{\includegraphics[width=0.96\linewidth]{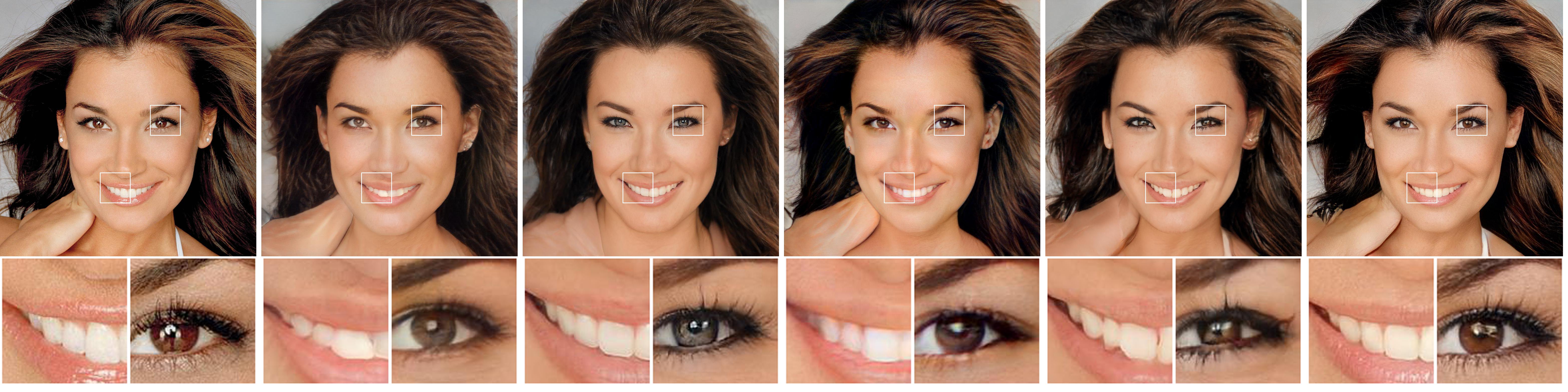}}
\\
\end{tabular} 
\renewcommand{\arraystretch}{1}
\begin{tabular}{P{0.16\linewidth}|P{0.08\linewidth}P{0.08\linewidth}P{0.08\linewidth}P{0.08\linewidth}P{0.08\linewidth}P{0.08\linewidth}P{0.08\linewidth}P{0.08\linewidth}P{0.08\linewidth}P{0.08\linewidth}P{0.08\linewidth}}
\centering
& Image & Patch  
& Image & Patch 
& Image & Patch 
& Image & Patch
& Image & Patch
\\
\hline
MSE $\downarrow$  & 0.012 & 0.013 & 0.019 & 0.014 & 0.012 & 0.009 & 0.009 & 0.011 & \textbf{0.008} & \textbf{0.009} \\
LPIPS $\downarrow$  & 0.152 & 0.350 & 0.203 & 0.301 & 0.117 & 0.323 & 0.111 & 0.324 & \textbf{0.066} & \textbf{0.201}
\\
\end{tabular} 
\renewcommand{\arraystretch}{0}
\begin{tabular}{P{0.16\linewidth}P{0.16\linewidth}P{0.16\linewidth}P{0.16\linewidth}P{0.16\linewidth}P{0.16\linewidth}}
\centering
&&&&&
\\
\multicolumn{6}{c}{\includegraphics[width=0.96\linewidth]{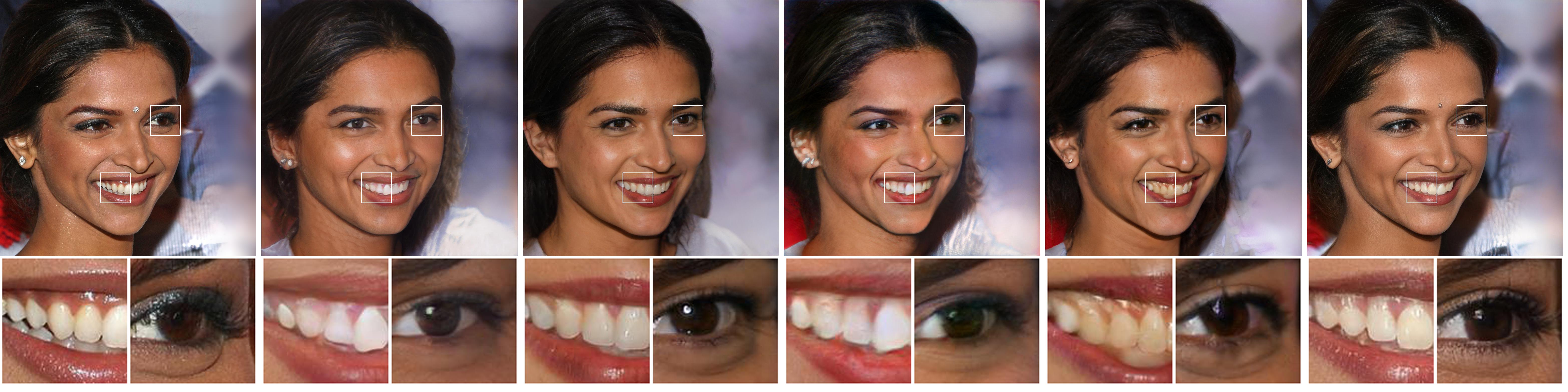}}
\\
\end{tabular} 
\renewcommand{\arraystretch}{1}
\begin{tabular}{P{0.16\linewidth}|P{0.08\linewidth}P{0.08\linewidth}P{0.08\linewidth}P{0.08\linewidth}P{0.08\linewidth}P{0.08\linewidth}P{0.08\linewidth}P{0.08\linewidth}P{0.08\linewidth}P{0.08\linewidth}P{0.08\linewidth}}
\centering
& Image & Patch  
& Image & Patch 
& Image & Patch 
& Image & Patch
& Image & Patch
\\
\hline
MSE $\downarrow$  & 0.009 & 0.013 & 0.011 & 0.016 & 0.006 & 0.012 & 0.005 & 0.016 & \textbf{0.003} & \textbf{0.011} \\
LPIPS $\downarrow$  & 0.145 & 0.291 & 0.189 & 0.262 & 0.117 & 0.310 & 0.112 & 0.320 & \textbf{0.054} & \textbf{0.194}
\end{tabular}  
\caption{\textbf{Inversion on face domain}. We compare our model against state-of-the-art encoder-based methods \cite{richardson2020encoding,tov2021designing,alaluf2021restyle,wang2021high} for the inversion of StyleGAN2 pre-trained on face domain. Reconstructions using our framework are visually more faithful and zoom-in patches show that they exhibit much more details and sharpness. Pixel-wise reconstruction errors (MSE error, lower is better) and perceptual quality (LPIPS distance, lower is better) confirm this visual conclusion on these examples. 
}
\label{compare_recon}
\end{figure*}
\subsection{Experimental setup} 

We evaluate our framework on several style-based generators pre-trained on various domains. We train the encoder for the inversion of StyleGAN2~\cite{karras2019analyzing} on faces and cars, and StyleGAN2-Ada~\cite{karras2020ada} on cats and dogs. %
For each generator pre-trained on a specific domain, a separate encoder is trained. During the training, we use a batch size of $4$, each batch containing two real images and two synthetic images. The model is trained for $12$ epochs, using $10K$ iterations per epoch. The learning rate is $10^{-4}$ for the first $10$ epochs and is divided by ten for the last $2$ epochs. For the face domain, we minimize Eq.(\ref{eq:final_loss}), using FFHQ~\cite{karras2019style} for training, and CelebA-HQ~\cite{karras2018progressive} for evaluation. For the car domain, we minimize Eq.(\ref{eq:total_loss}), using Stanford Cars~\cite{krause20133d} training set for training, and the corresponding test set for evaluation. For the cat/dog domain, we minimize Eq.(\ref{eq:total_loss}), using AFHQ Cats/Dogs~\cite{choi2020stargan} train set for training, and the corresponding test set for evaluation.

\subsection{Inversion}

We evaluate our model against the current state-of-the-art encoder-based GAN inversion methods: pSp~\cite{richardson2020encoding}, e4e~\cite{tov2021designing}, restyle~\cite{alaluf2021restyle} and a recent preprint work HFGI~\cite{wang2021high}. We first perform comparisons for the inversion of StyleGAN2 model pre-trained on FFHQ dataset. For each method we use the official implementation~\cite{codepsp,codee4e,coderestyle,codeHFGI} to generate the results. Restyle has been implemented on both pSp and e4e. We use restyle-pSp as it has better performance.

\noindent \textbf{Qualitative Results} \quad Figure~\ref{compare_recon} shows the inversion results of the different methods. Overall, visual inspection shows that our method outperforms other models. Firstly, faces are more faithfully reconstructed globally. Secondly, zoom-in patches show that more details are preserved and that the images produced by our framework are significantly sharper than those of the concurrent methods.

\begin{table}[t]
\footnotesize
\centering
\setlength{\tabcolsep}{0pt}
\renewcommand{\arraystretch}{1.1}
\begin{tabular}{|P{0.19\linewidth}|P{0.13\linewidth}|P{0.13\linewidth}|P{0.13\linewidth}|P{0.13\linewidth}|P{0.13\linewidth}|P{0.13\linewidth}|}
\hline
Method & SSIM $\uparrow$ & PSNR $\uparrow$ & MSE $\downarrow$ & LPIPS $\downarrow$ & ID $\uparrow$ & FID $\downarrow$  
\\
\hline
IDGI\cite{zhu2020domain} & 0.554 & 22.06 & 0.034 & 0.136 & 0.480 & 36.83 \\
pSp\cite{richardson2020encoding}  & 0.509 & 20.37 & 0.040 & 0.159 & 0.654 & 34.68  \\
e4e\cite{tov2021designing} & 0.479 & 19.17 & 0.052 & 0.196 & 0.593 & 36.72  \\
restyle\cite{alaluf2021restyle} & 0.537 & 21.14 & 0.034 & 0.130 & 0.735 & 32.56  \\
HFGI\cite{wang2021high} & 0.595 & 22.07 & 0.027 &  0.117 & 0.758  & 26.53  \\
Ours & \textbf{0.641} & \textbf{23.65} & \textbf{0.019} & \textbf{0.066}  & \textbf{0.867}& \textbf{19.03}  
\\
\hline
\end{tabular}
\caption{\textbf{Quantitative evaluation.}
We use \textit{SSIM}, \textit{PSNR} and \textit{MSE} to measure the reconstruction error, and \textit{LPIPS}\cite{zhang2018unreasonable} for the perceptual quality.
We also measure the \textit{identity similarity} (ID) between the inversion and the source image, which refers to the cosine similarity between the features in ArcFace \cite{deng2019arcface} of both images.  
To measure the discrepancy between the real data distribution and the inversion one, we use \textit{FID} \cite{heusel2017gans}. 
Overall, our method outperforms the state-of-the-art baselines by up to $10\% - 20\%$. In terms of perceptual quality (LPIPS), we improve the result by $50\%$.}
\label{quan_table}
\end{table}

\begin{figure}[t]
\centering
\small
\setlength{\tabcolsep}{0pt}
\renewcommand{\arraystretch}{0}
\begin{tabular}{P{0.23\linewidth}P{0.23\linewidth}P{0.23\linewidth}P{0.23\linewidth}}
Source & e4e \cite{tov2021designing} & restyle-e4e \cite{alaluf2021restyle} & Ours
\\
\multicolumn{4}{c}{\includegraphics[width=0.92\linewidth]{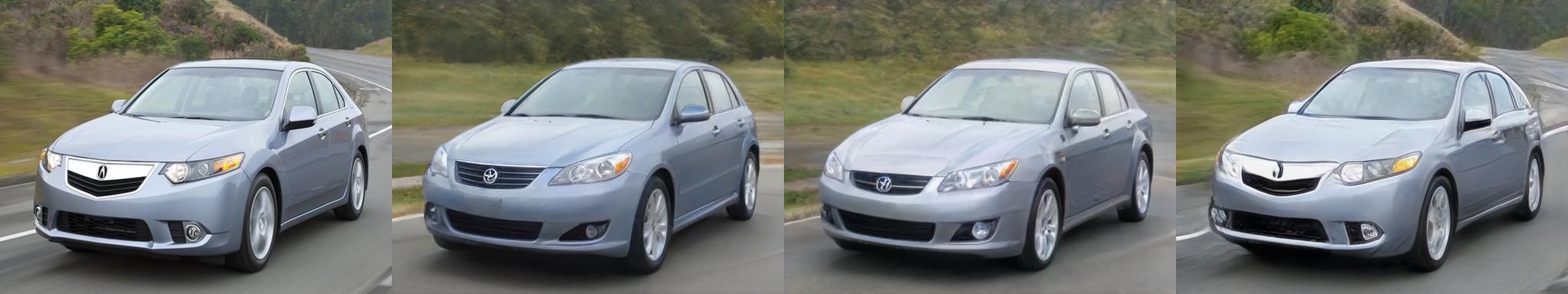}}
\\
\multicolumn{4}{c}{\includegraphics[width=0.92\linewidth]{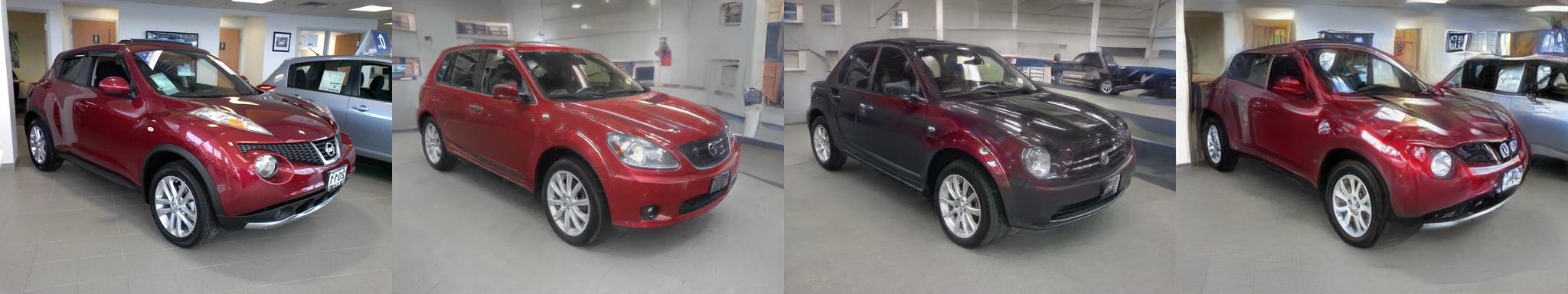}}
\\
\end{tabular} 
\\
(a) Inversion on car domain.
\vspace{2pt}
\begin{tabular}{P{0.23\linewidth}P{0.23\linewidth}P{0.23\linewidth}P{0.23\linewidth}}
Source & Inversion & Source & Inversion 
\vspace{2pt}
\\
\multicolumn{4}{c}{\includegraphics[width=0.92\linewidth]{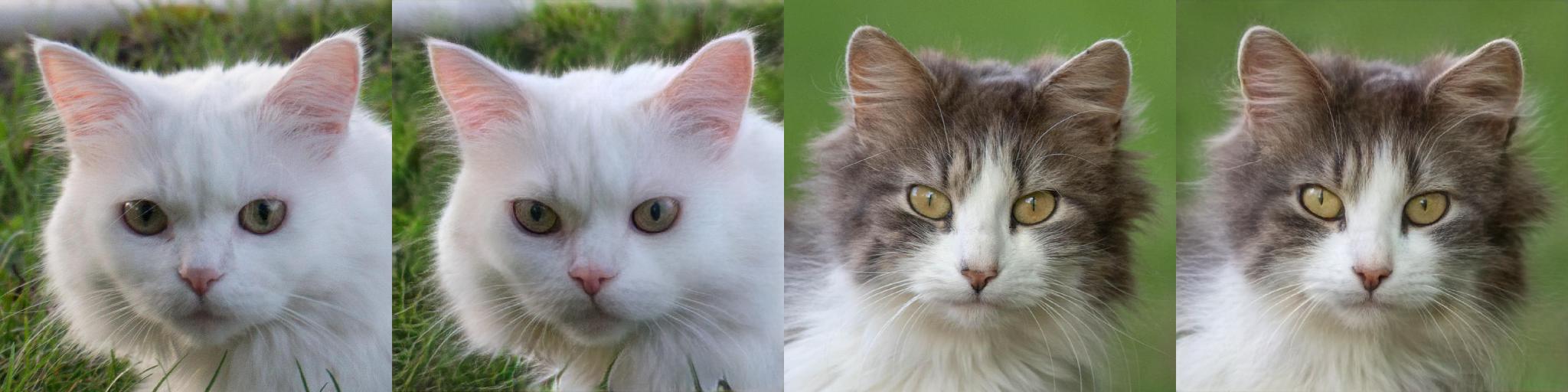}}
\\
\multicolumn{4}{c}{\includegraphics[width=0.92\linewidth]{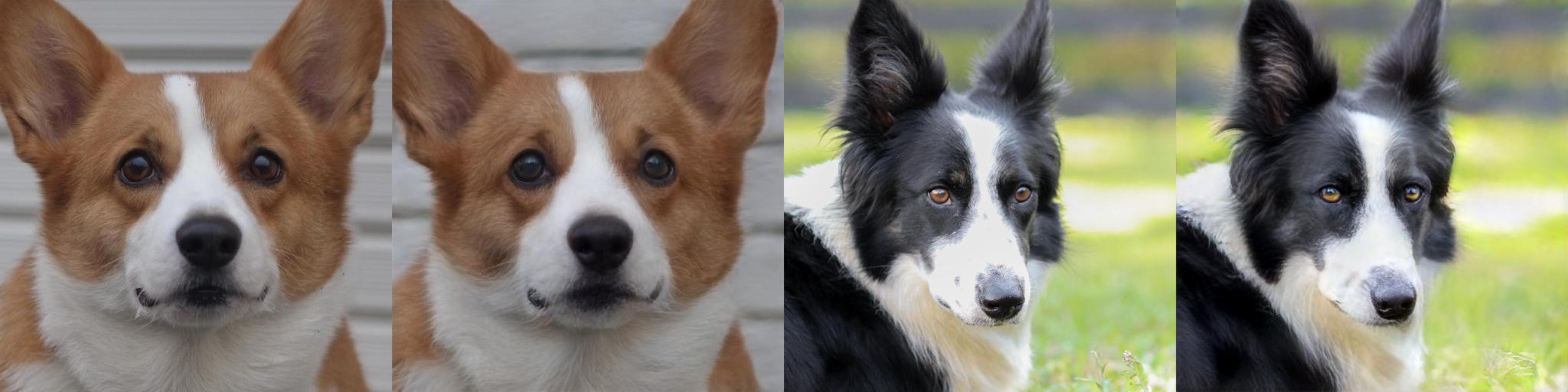}}
\\
\end{tabular} 
\\
(b) Inversion on cat/dog domain.
\caption{\textbf{Inversion on other domains.} In (a), we show the inversion results of StyleGAN2 pre-trained on car domain. Our method captures better the details than e4e \cite{tov2021designing} and restyle-e4e \cite{alaluf2021restyle}.
In (b), we show the inversion results of StyleGAN2-Ada pre-trained on the cat and dog domains, respectively.
}
\label{inversion_other}
\end{figure}

\noindent \textbf{Quantitative Evaluation} \quad 
We evaluate our approach quantitatively against the aforementioned encoder based methods~\cite{richardson2020encoding,tov2021designing,alaluf2021restyle,wang2021high} and a hybrid method (in-domain GAN)~\cite{zhu2020domain}.
We compare each method on the inversion of StyleGAN2 pretrained on FFHQ, using the first $1K$ images of CelebA-HQ as evaluation data. 
To measure the reconstruction error, we compute \emph{SSIM}, \emph{PSNR} and \emph{MSE}. 
To measure the perceptual quality, we measure the \emph{LPIPS}\cite{zhang2018unreasonable} distance.
Additionally, we measure the \emph{identity similarity} (ID) between the inversion and the source image, which refers to the cosine similarity between the features in ArcFace~\cite{deng2019arcface} of the two images.  
To measure the discrepancy between the real data distribution and the inversion one, we use the Frechet Inception Distance~\cite{heusel2017gans} (\emph{FID}). 
Table~\ref{quan_table} presents the quantitative evaluation of all the methods. Our method significantly outperforms the state-of-the-art methods on all the metrics. In terms of perceptual quality (LPIPS), improvement can attain $50\%$.  

\noindent \textbf{Inversion for other domains} \quad 
Figure~\ref{inversion_other}(a) shows the inversion for StyleGAN2 pretrained on the car domain. We train the encoder with Stanford Car dataset~\cite{krause20133d}. Compared with e4e~\cite{tov2021designing} and restyle-e4e\cite{alaluf2021restyle}, our inversion achieves a better reconstruction quality, preserving better the details. Figure~\ref{inversion_other}(b) shows the inversion for StyleGAN2-Ada pretrained on AFHQ Cat/Dog dataset~\cite{choi2020stargan}. Our encoder achieves nearly perfect inversions. Here we did not compare with~\cite{tov2021designing,alaluf2021restyle}, as the official pre-trained model is unavailable.

\begin{figure}[t]
\centering
\small
\setlength{\tabcolsep}{0pt}
\renewcommand{\arraystretch}{0.5}
\begin{tabular}{P{0.04\linewidth}P{0.18\linewidth}P{0.18\linewidth}P{0.18\linewidth}P{0.18\linewidth}P{0.18\linewidth}}
\centering
& pSp\cite{richardson2020encoding} 
& e4e\cite{tov2021designing} 
& restyle \cite{alaluf2021restyle} 
& HFGI\cite{wang2021high} 
& Ours
\\
\rotatebox{90}{
\begin{tabular}{*{5}{>{\centering\arraybackslash}m{42pt}}}
Eyeglasses 
& Smiling 
& Makeup 
& Inversion 
& Source \\
\end{tabular}
}
&
\includegraphics[width=\linewidth]{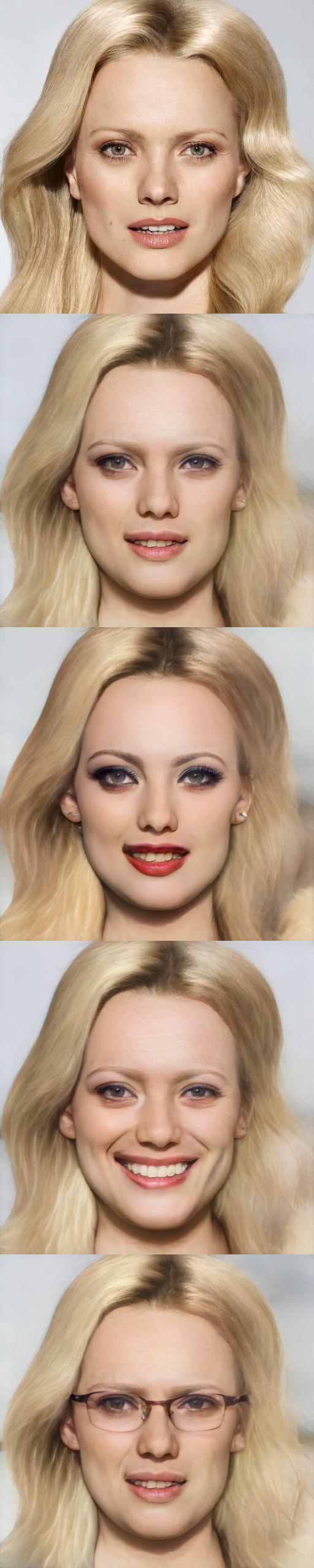}
& \includegraphics[width=\linewidth]{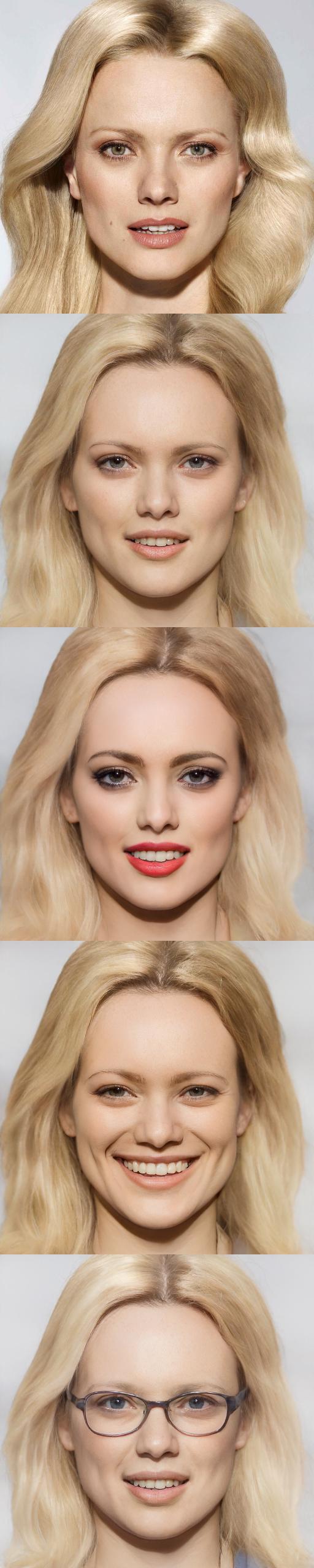}
& \includegraphics[width=\linewidth]{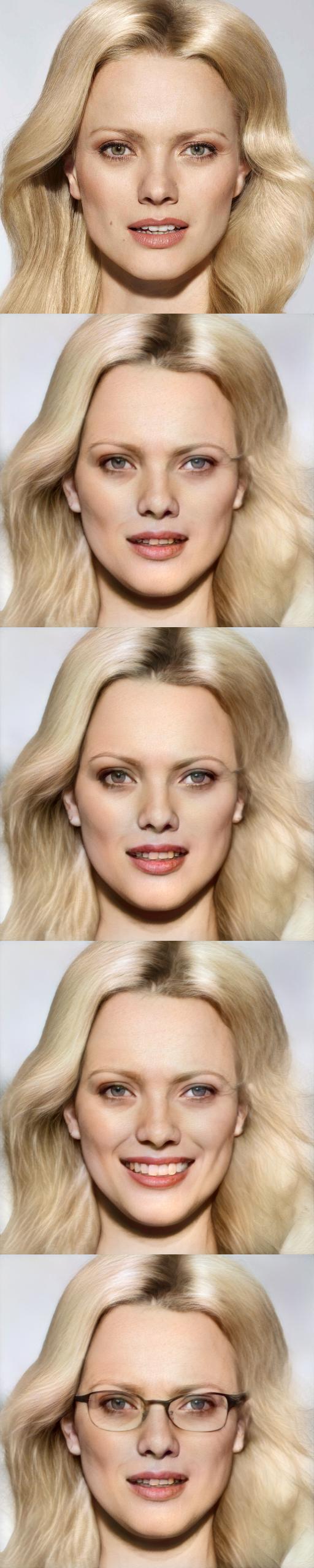}
& \includegraphics[width=\linewidth]{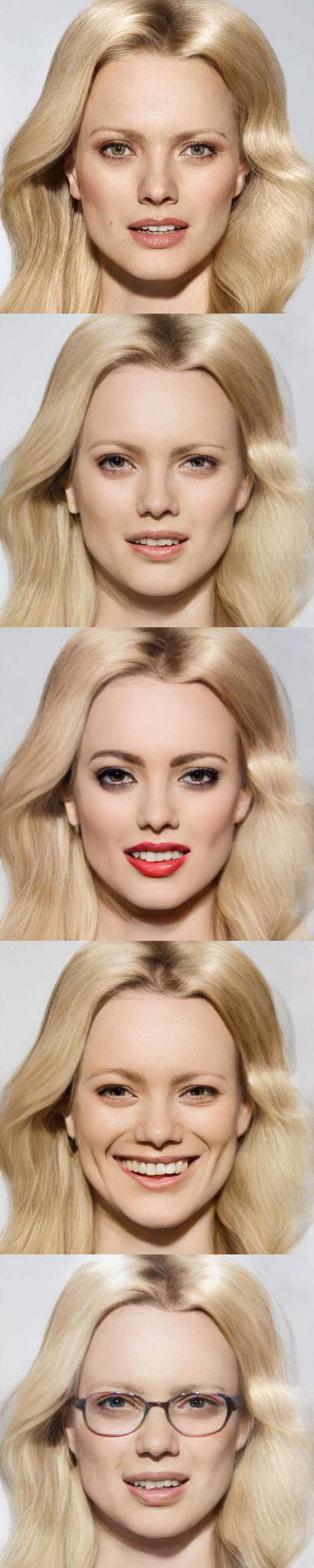}
& \includegraphics[width=\linewidth]{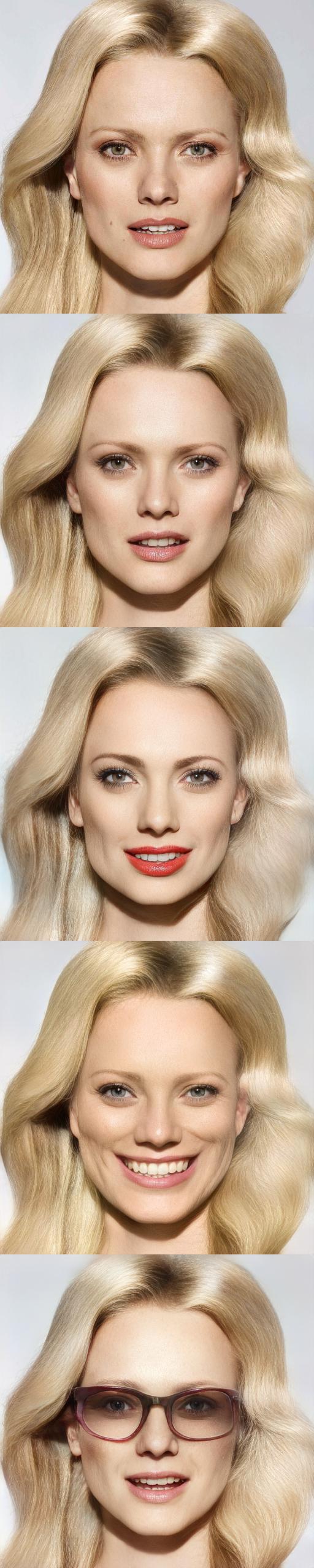}
\end{tabular} 
\caption{\textbf{Latent space editing}. For each method, we apply InterFaceGAN \cite{shen2019interpreting} to perform latent editing for facial attribute manipulation. Our method yields plausible editing results, while at the same time preserving better the identity and the sharpness.%
}
\label{compare_edit}
\end{figure}

\subsection{Editing}

In this experiment, we consider the task of real image editing via latent space manipulation. We compare our approach with the state-of-the-art encoder-based GAN inversion methods\cite{richardson2020encoding,tov2021designing,alaluf2021restyle,wang2021high} on the facial image editing via the latent space of StyleGAN2 pretrained on FFHQ dataset. Despite the fact that all the encoders project real images to the latent space of StyleGAN, the latent distribution learned by each encoder can be different. As such, for each inversion model, we generate the inverted latent codes for the first $10K$ images of CelebA-HQ, and apply InterFaceGAN~\cite{shen2019interpreting} to find the editing directions in the learned latent space. Figure~\ref{compare_edit} shows facial attribute editing results for all methods. Compared with the state-of-the-art, our method yields visually plausible editing results, while preserving better the identity and sharpness. 

\begin{table}[t]
\footnotesize
\centering
\setlength{\tabcolsep}{0pt}
\renewcommand{\arraystretch}{1.1}
\begin{tabular}{|P{0.16\linewidth}|P{0.16\linewidth}|P{0.16\linewidth}|P{0.16\linewidth}|P{0.16\linewidth}|P{0.16\linewidth}|}
\hline
Method 
& pSp\cite{richardson2020encoding} 
& e4e\cite{tov2021designing}  
& restyle \cite{alaluf2021restyle} 
& HFGI\cite{wang2021high} 
& Ours 
\\
\hline
FID$\downarrow$  & 39.66 &  37.79 & 32.84 & 30.68 &\textbf{27.45}
\\
\hline
\end{tabular}
\caption{\textbf{FID measured on edited images.} For each inversion method, we encode the first $1K$ images of CelebA-HQ to the latent space and perform latent editing on four facial attributes. The FID is calculated between the real images and all the edited images. Our method outperforms the other approaches. }
\label{editing_fid}
\end{table}

To evaluate quantitatively the editing results of each method, we take the first $1K$ images of CelebA-HQ as testing data. For each method, we project each image to the latent space and apply InterFaceGAN~\cite{shen2019interpreting} to generate the editing result on the following facial attributes: `gender', `makeup', `smiling' and `eyeglasses'. Then we compute the FID between the real images and all the edited images. Table~\ref{editing_fid} shows that the discrepancy between the data distribution of our editing results and that of the real images is the smallest by a large margin.

\begin{figure}[t]
\centering
\small
\setlength{\tabcolsep}{0pt}
\renewcommand{\arraystretch}{0.5}
\begin{tabular}{P{0.24\linewidth}P{0.24\linewidth}P{0.24\linewidth}P{0.24\linewidth}}
\centering
$\ve{G}(\ve{w}_A,\ve{F}_A)$ & $\ve{G}(\ve{w}_B,\ve{F}_A)$ & $\ve{G}(\ve{w}_A,\ve{F}_B)$ & $\ve{G}(\ve{w}_B,\ve{F}_B)$
\\
\multicolumn{4}{c}{\includegraphics[width=0.95\linewidth]{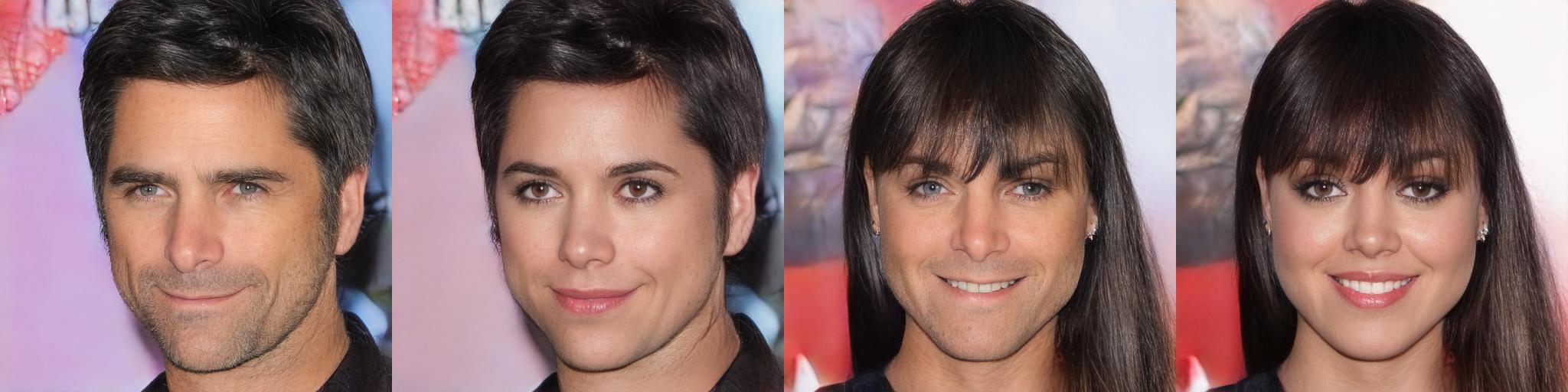}}
\\
\end{tabular} 
\caption{\textbf{Style mixing.} 
The first and last column show the inversions of two images $\ve{x}_A$ and $\ve{x}_B$, denoted by $\ve{G}(\ve{w}_A,\ve{F}_A)$ and $\ve{G}(\ve{w}_B,\ve{F}_B)$, respectively. The second column is generated from the feature code of $\ve{x}_A$ and the latent code of $\ve{x}_B$, denoted by $\ve{G}(\ve{w}_B,\ve{F}_A)$, and vice versa for the third column, denoted by $\ve{G}(\ve{w}_A,\ve{F}_B)$. The feature code encodes the geometric structures such as pose and facial shape, whereas the latent code controls the appearance styles like eye color and makeup.
}
\label{style_mixing}
\end{figure}

Additionally, we show style mixing results in Figure~\ref{style_mixing}, generated from the latent code of one image with the feature code of another image. From this experiment we observe that the geometric structures such as pose and facial shape are encoded by the feature code, while the appearance styles like eye color and makeup are encoded by the latent code.

\begin{figure}[t]
\centering
\small
\setlength{\tabcolsep}{0.3pt}
\renewcommand{\arraystretch}{0}
\begin{tabular}{P{0.04\linewidth}P{0.9\linewidth}}
\rotatebox{90}{\quad Source}
&\includegraphics[width=\linewidth]{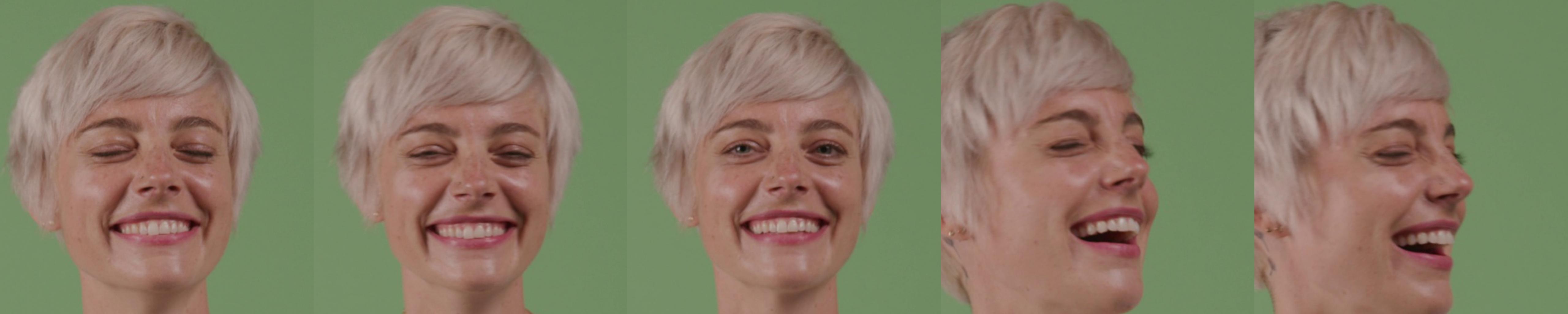}
\\
\rotatebox{90}{\quad pSp\cite{richardson2020encoding} }
&\includegraphics[width=\linewidth]{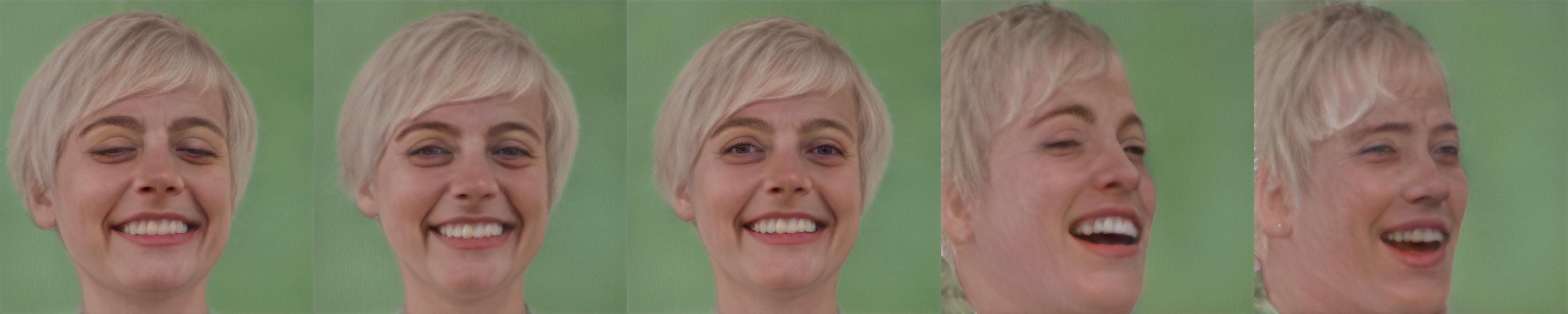}
\\
\rotatebox{90}{\quad e4e\cite{tov2021designing} }
&\includegraphics[width=\linewidth]{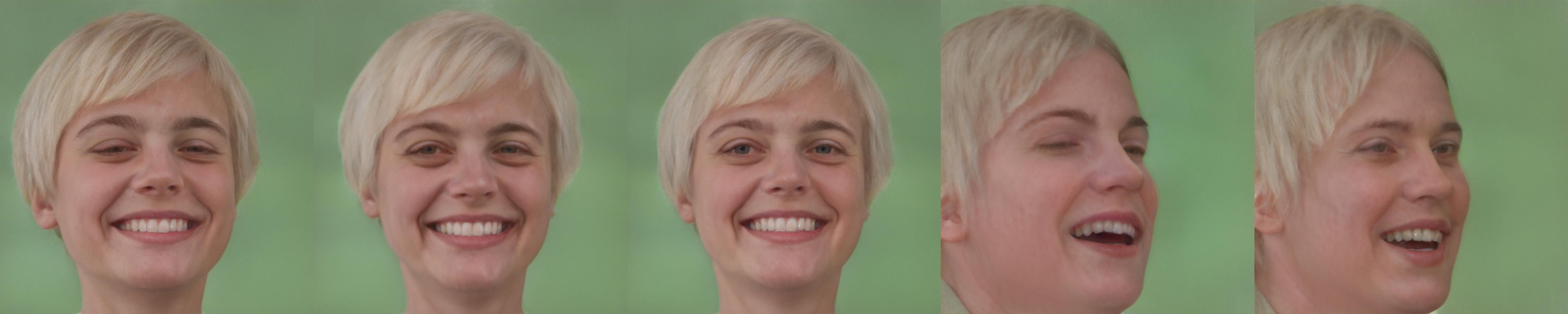}
\\
\rotatebox{90}{restyle\cite{alaluf2021restyle} }
&\includegraphics[width=\linewidth]{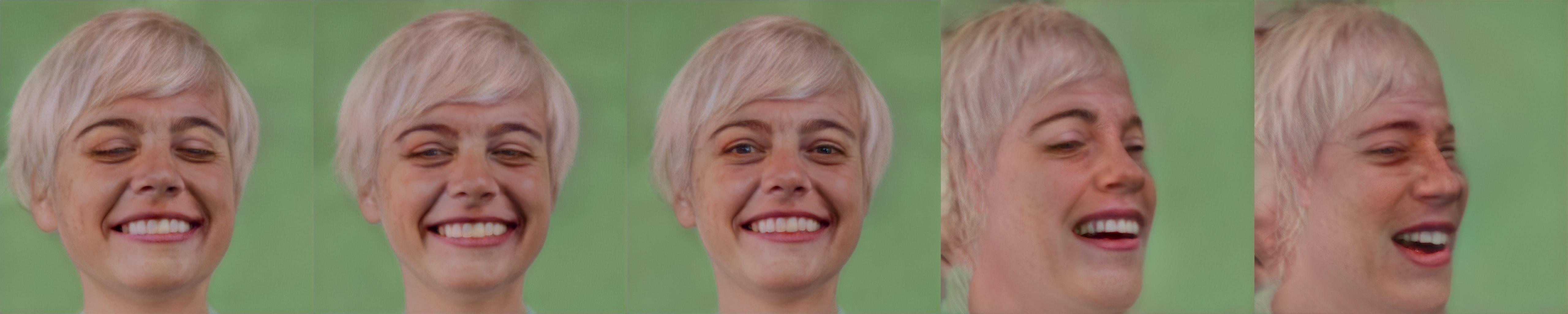}
\\
\rotatebox{90}{HFGI\cite{wang2021high} }
&\includegraphics[width=\linewidth]{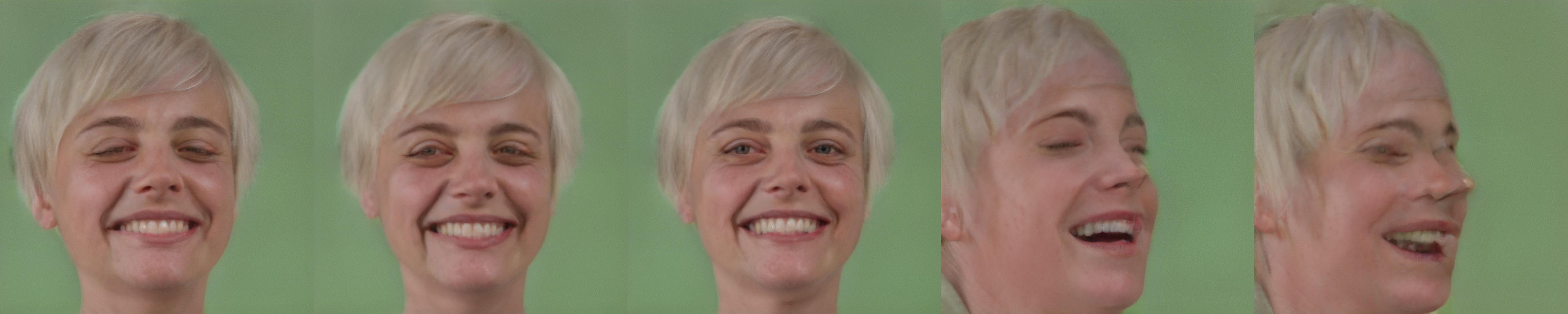}
\\
\raisebox{1.5\normalbaselineskip}[0pt][0pt]{\rotatebox[origin=c]{90}{Ours}}
&\includegraphics[width=\linewidth]{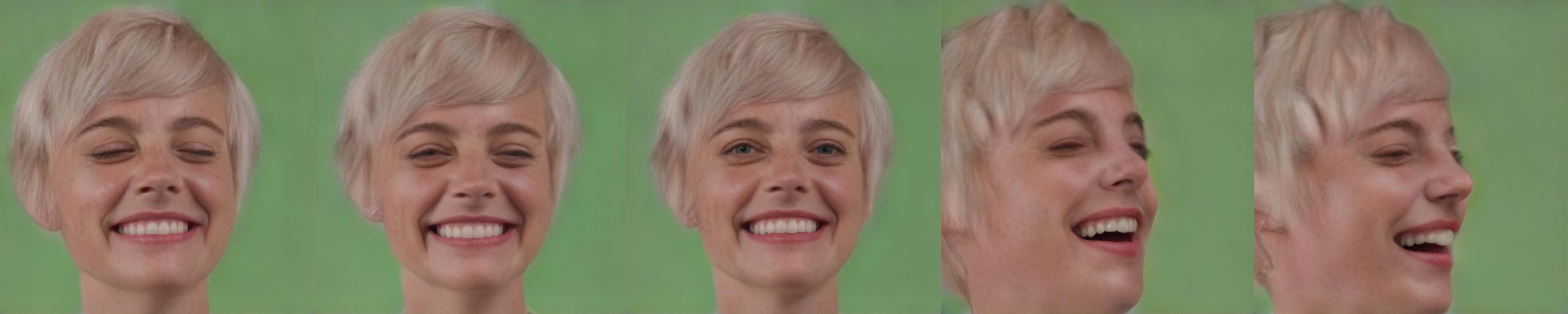}
\\
\end{tabular}
\caption{\textbf{Video inversion}. For each method, we show the inversion results of multiple consecutive images extracted from a video sequence. Our inversion method preserves better the identity along the video and yields a better reconstruction for the extreme poses.
}
\label{compare_video}
\end{figure}

\subsection{Video inversion}

In this section, we discuss the possibility of integrating our proposed encoder into a video editing pipeline. We compare the inversion quality and stability of different encoders on videos. Figure~\ref{compare_video} shows a qualitative inversion result on consecutive images extracted from a video sequence. The last two frames in the original sequence are extreme poses. As can be observed, other methods fail to invert non-frontal poses, thus damaging the consistent reconstruction along the sequence. Our approach yields consistent inversion of high fidelity, which favors further editing on videos.

\begin{table}[t]
\footnotesize
\centering
\setlength{\tabcolsep}{0pt}
\renewcommand{\arraystretch}{1.1}
\begin{tabular}{|P{0.25\linewidth}|P{0.14\linewidth}|P{0.14\linewidth}|P{0.14\linewidth}|P{0.14\linewidth}|P{0.14\linewidth}|}
\hline
Method  & SSIM $\uparrow$ & PSNR $\uparrow$ & MSE $\downarrow$ & LPIPS $\downarrow$ & ID $\uparrow$
\\
\hline
pSp\cite{richardson2020encoding} & 0.736 & 22.30 & 0.025 & 0.196 & 0.687 \\
e4e\cite{tov2021designing}& 0.713 & 20.57 & 0.037 & 0.220 & 0.620 \\
restyle-pSp \cite{alaluf2021restyle}& 0.761 & 23.17 & 0.021 & 0.189 & 0.781 \\
HFGI \cite{wang2021high}& 0.783 & 24.04 & 0.017 & 0.182 & 0.810\\
Ours & \textbf{0.818} & \textbf{26.64} & \textbf{0.009} & \textbf{0.122} & \textbf{0.895} \\
\hline
\end{tabular}
\caption{\textbf{Quantitative evaluation on video inversion.} 
We sample randomly 120 videos from RAVDESS dataset\cite{livingstone2018ryerson}, perform the inversion using each method and compute the quantitative metrics. Our method outperforms the competing approaches on both the reconstruction error and the perceptual quality.
}
\label{quan_video}
\end{table}

We evaluate our encoder quantitatively against the state-of-the-art for video inversion on RAVDESS~\cite{livingstone2018ryerson}, a dataset of talking face videos. From which we sample randomly 120 videos as evaluation data. For each method, we perform the inversion on each video and compute the quantitative metrics on the inversion results. As shown in Table~\ref{quan_video}, our approach outperforms the competing approaches on both the reconstruction error and the perceptual quality.

\begin{table}[t]
\footnotesize
\centering
\setlength{\tabcolsep}{0pt}
\renewcommand{\arraystretch}{1.1}
\begin{tabular}{|L{0.31\linewidth}|P{0.13\linewidth}|P{0.13\linewidth}|P{0.13\linewidth}|P{0.12\linewidth}|P{0.13\linewidth}|}
\hline
\ Configurations 
& SSIM $\uparrow$ 
& PSNR $\uparrow$ 
& LPIPS $\downarrow$ 
& ID $\uparrow$
& FID $\downarrow$ 
\\
\hline
\ (A) w/o $\mathcal{L}_{m\_lpips}$ & 0.647 & 24.01 & 0.056 & 0.874
& 22.63 
\\
\ (B) w/o feature input & 0.489 & 19.44 &  0.192 & 0.635 
& 35.28 
\\
\ (C) w/o synthetic data & 0.644 & 23.67 & 0.065 & 0.873
& 20.45 
\\
\ (D) our baseline & 0.641 & 23.65 & 0.066 & 0.867
& 19.03
\\
\hline
\end{tabular}
\caption{
\textbf{Ablative study on experimental setup.} We conduct experiments on three different configurations: (1) w/o multi-scale setting in the perceptual loss, (2) w/o feature prediction branch, (3) w/o synthetic training data. Our baseline achieves better perceptual quality and comparable performance on the distortion metrics. 
}
\label{config}
\end{table}
\subsection{Ablation study}
We conduct an ablation study on the experimental setup for the inversion of StyleGAN2 pretrained on FFHQ. Specifically, we compare the quantitative metrics of several ablative configurations. As shown in Table~\ref{config}, in configuration (A), we replace the multi-scale perceptual loss by a common LPIPS loss, and change the corresponding weight $\lambda_1$ to scale it to a similar magnitude as before. In (B), we discard the feature prediction branch and generate the inversion with only the latent code. In (C), we use only real images as training data.

For (A), we observe a comparable result on the distortion metrics, but a much higher FID compared to the baseline. Including the proposed multi-scale perceptual loss greatly improves the perceptual quality and presents comparable performance on other distortion metrics. (B) confirms that the feature code helps to generate an inversion with better fidelity. For (C), we observe similar performance on the distortion metrics but a higher FID value. This demonstrates that including synthetic data in the training helps improving the perceptual quality of the inversion results.

\section{Conclusion}
\label{sec:discussion}

We have proposed a new \emph{Feature-Style encoder} architecture for style-based GAN inversion. For the first time, we achieve projection of a given image to the Feature-Style latent space of a style-based generator, without out any optimization step at inference time. Our approach significantly improves the perceptual quality of the inversion, outperforming the strong and competitive state-of-the-art methods. Additionally, we show that our proposed encoder is more suited for the inversion and editing of videos.

{\small
\bibliographystyle{ieee_fullname}
\bibliography{egbib}
}

\clearpage
\appendix



\section{Inversion}

\begin{figure*}[t]
\centering
\small
\setlength{\tabcolsep}{0pt}
\renewcommand{\arraystretch}{1.0}
\begin{tabular}{P{0.25\linewidth}P{0.25\linewidth}P{0.25\linewidth}P{0.25\linewidth}}
\centering
Source & Optimization~\cite{abdal2020image2stylegan++} & In-domain~\cite{zhu2020domain}& pSp~\cite{richardson2020encoding} 
\\
\multicolumn{4}{c}{\includegraphics[width=\linewidth]{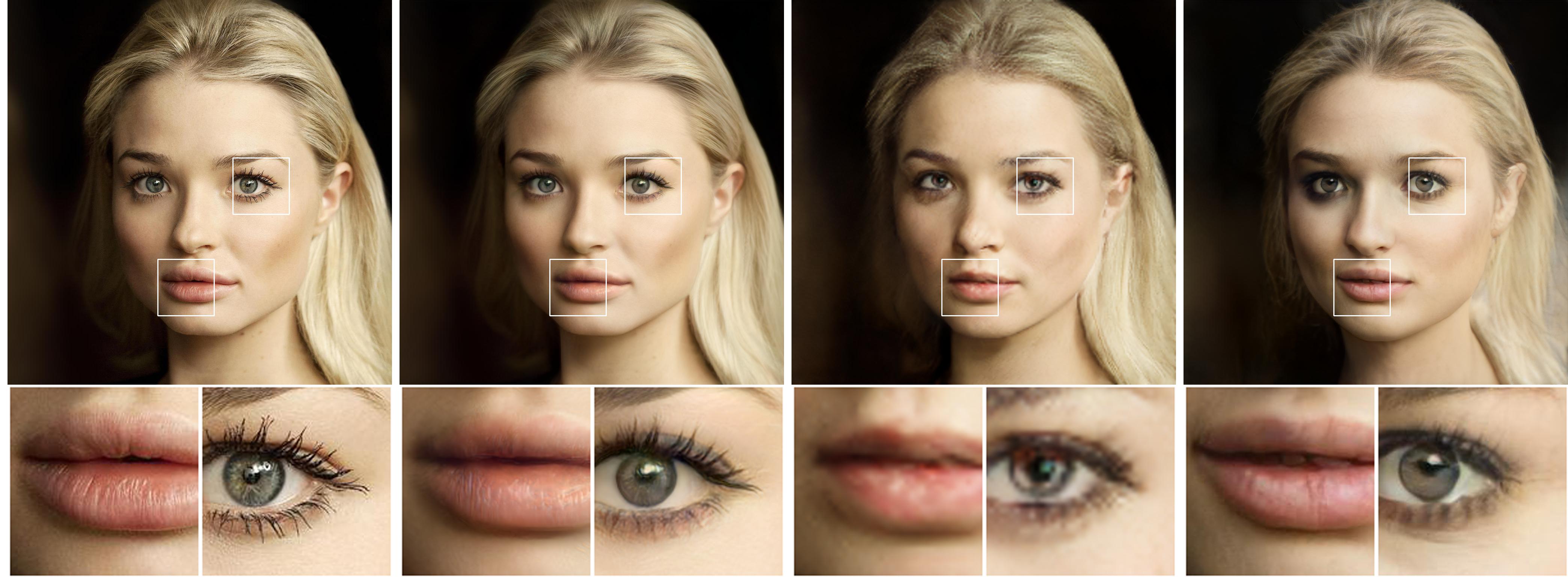}}
\\
e4e~\cite{tov2021designing} & restyle-pSp~\cite{alaluf2021restyle} & HFGI~\cite{wang2021high} & Ours
\\
\multicolumn{4}{c}{\includegraphics[width=\linewidth]{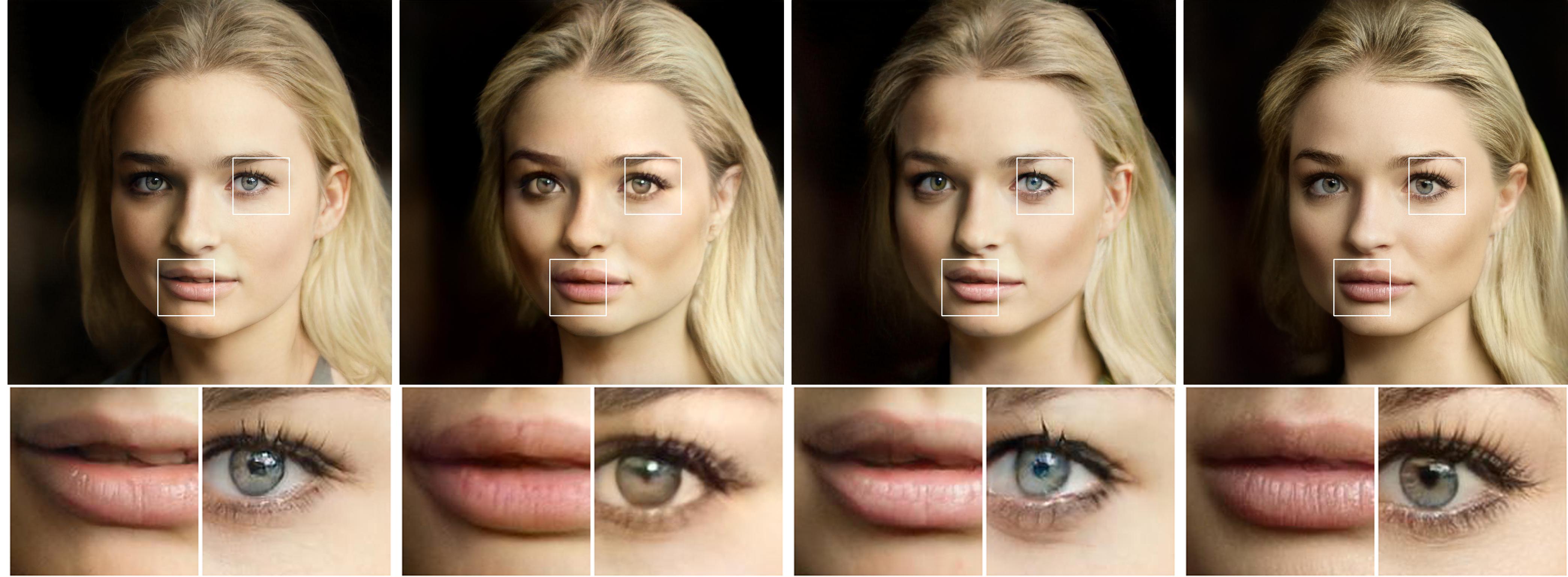}}
\\
\end{tabular}
\caption{\textbf{Inversion on face domain}. We compare our model against state-of-the-art GAN inversion methods~\cite{abdal2020image2stylegan++,zhu2020domain,richardson2020encoding,tov2021designing,alaluf2021restyle,wang2021high} for the inversion of StyleGAN2 pre-trained on face domain. Reconstructions using our framework are visually more faithful and zoom-in patches show that they exhibit much more details and sharpness.
}
\label{inversion_face}
\end{figure*}

\begin{figure*}[t]
\centering
\small
\setlength{\tabcolsep}{0pt}
\renewcommand{\arraystretch}{1.0}
\begin{tabular}{P{0.25\linewidth}P{0.25\linewidth}P{0.25\linewidth}P{0.25\linewidth}}
\centering
Source & Optimization~\cite{abdal2020image2stylegan++} & In-domain~\cite{zhu2020domain}& pSp~\cite{richardson2020encoding} 
\\
\multicolumn{4}{c}{\includegraphics[width=\linewidth]{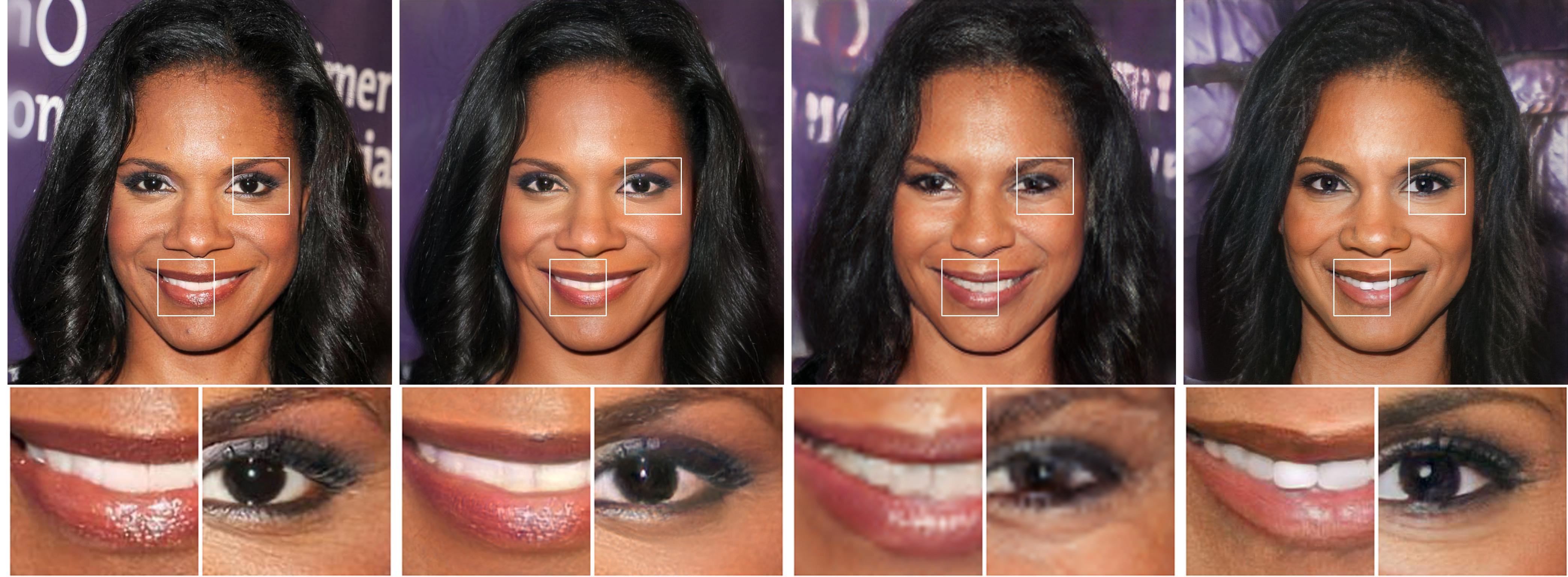}}
\\
e4e~\cite{tov2021designing} & restyle-pSp~\cite{alaluf2021restyle} & HFGI~\cite{wang2021high} & Ours
\\
\multicolumn{4}{c}{\includegraphics[width=\linewidth]{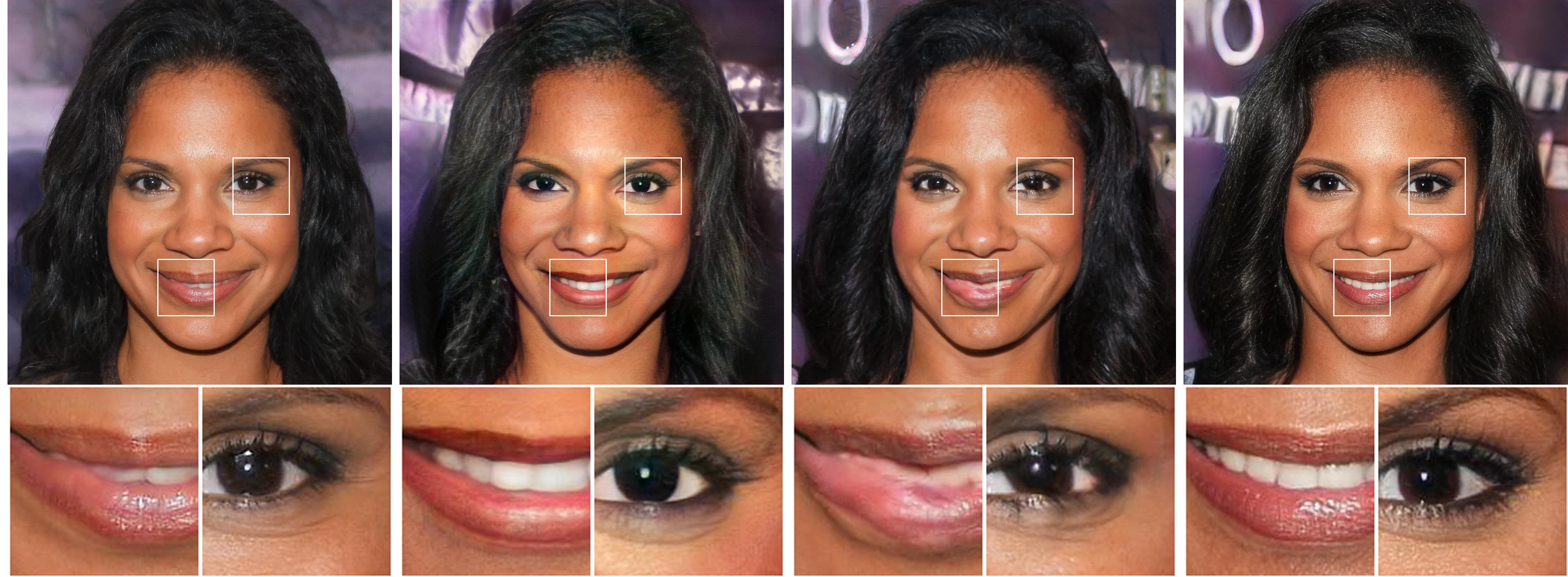}}
\\
\end{tabular}
\caption{\textbf{Inversion on face domain}. We compare our model against state-of-the-art GAN inversion methods~\cite{abdal2020image2stylegan++,zhu2020domain,richardson2020encoding,tov2021designing,alaluf2021restyle,wang2021high} for the inversion of StyleGAN2 pre-trained on face domain. Reconstructions using our framework are visually more faithful and zoom-in patches show that they exhibit much more details and sharpness.
}
\label{inversion_face_2}
\end{figure*}

\begin{figure*}[t]
\centering
\small
\setlength{\tabcolsep}{0pt}
\renewcommand{\arraystretch}{1.0}
\begin{tabular}{P{0.25\linewidth}P{0.25\linewidth}P{0.25\linewidth}P{0.25\linewidth}}
\centering
Source & Optimization~\cite{abdal2020image2stylegan++} & In-domain~\cite{zhu2020domain}& pSp~\cite{richardson2020encoding} 
\\
\multicolumn{4}{c}{\includegraphics[width=\linewidth]{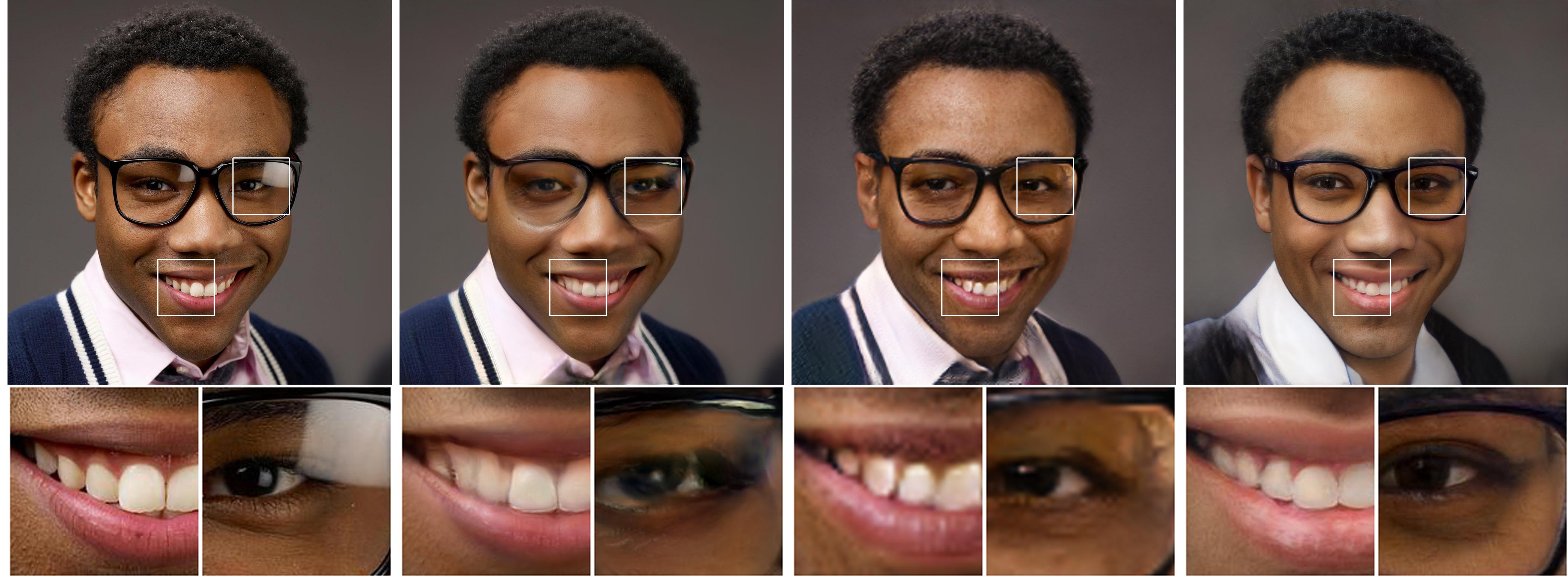}}
\\
e4e~\cite{tov2021designing} & restyle-pSp~\cite{alaluf2021restyle} & HFGI~\cite{wang2021high} & Ours
\\
\multicolumn{4}{c}{\includegraphics[width=\linewidth]{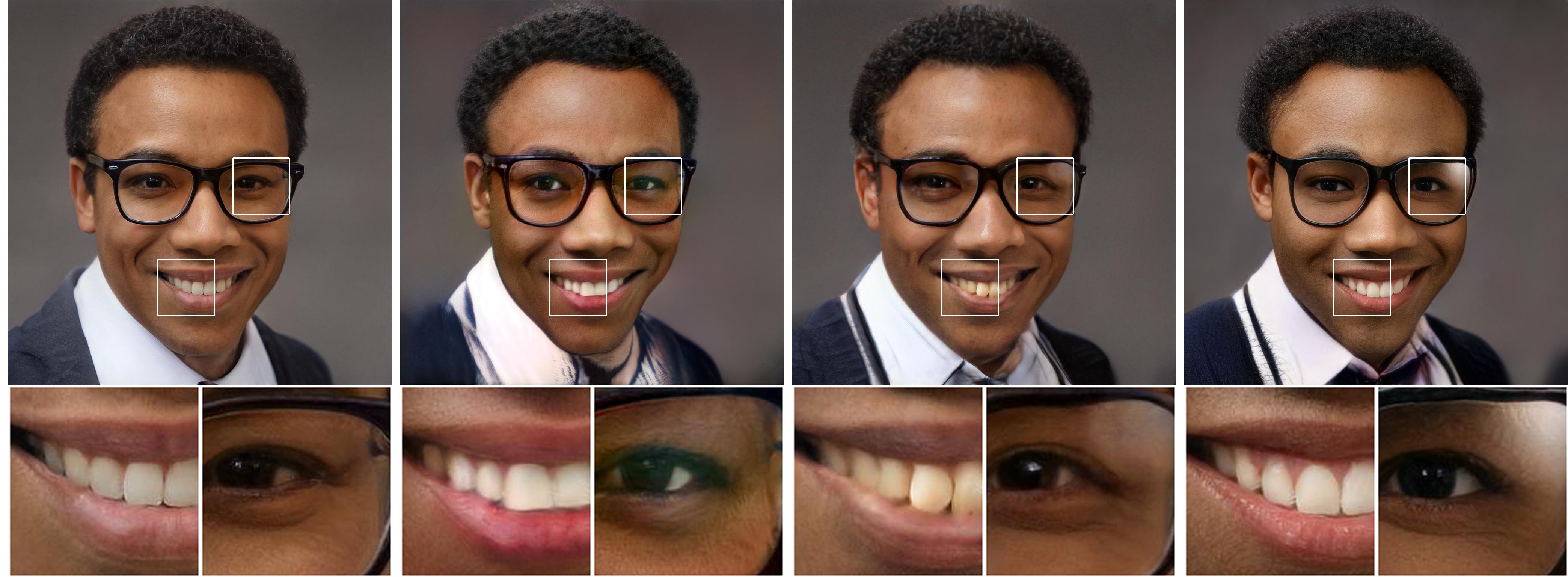}}
\\
\end{tabular}
\caption{\textbf{Inversion on face domain}. We compare our model against state-of-the-art GAN inversion methods~\cite{abdal2020image2stylegan++,zhu2020domain,richardson2020encoding,tov2021designing,alaluf2021restyle,wang2021high} for the inversion of StyleGAN2 pre-trained on face domain. Reconstructions using our framework are visually more faithful and zoom-in patches show that they exhibit much more details and sharpness.
}
\label{inversion_face_3}
\end{figure*}

\begin{figure*}[t]
\centering
\small
\setlength{\tabcolsep}{0pt}
\renewcommand{\arraystretch}{1.0}
\begin{tabular}{P{0.25\linewidth}P{0.25\linewidth}P{0.25\linewidth}P{0.25\linewidth}}
\centering
Source & Optimization~\cite{abdal2020image2stylegan++} & In-domain~\cite{zhu2020domain}& pSp~\cite{richardson2020encoding} 
\\
\multicolumn{4}{c}{\includegraphics[width=\linewidth]{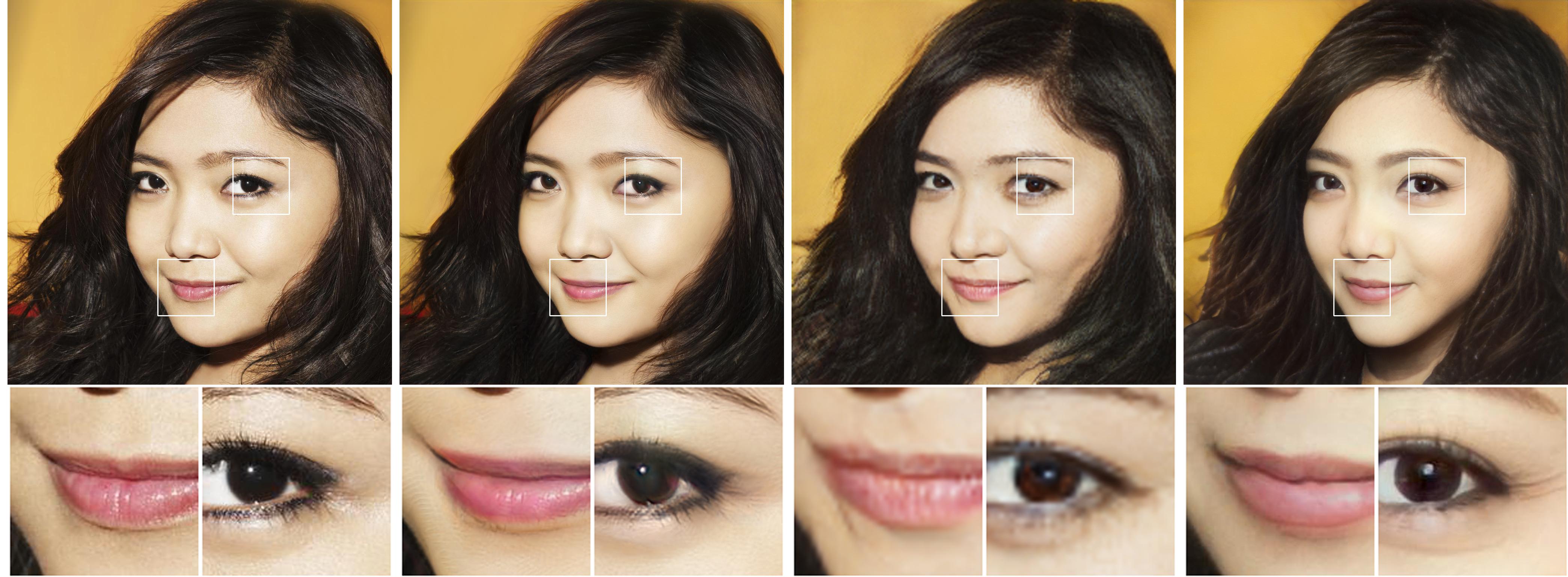}}
\\
e4e~\cite{tov2021designing} & restyle-pSp~\cite{alaluf2021restyle} & HFGI~\cite{wang2021high} & Ours
\\
\multicolumn{4}{c}{\includegraphics[width=\linewidth]{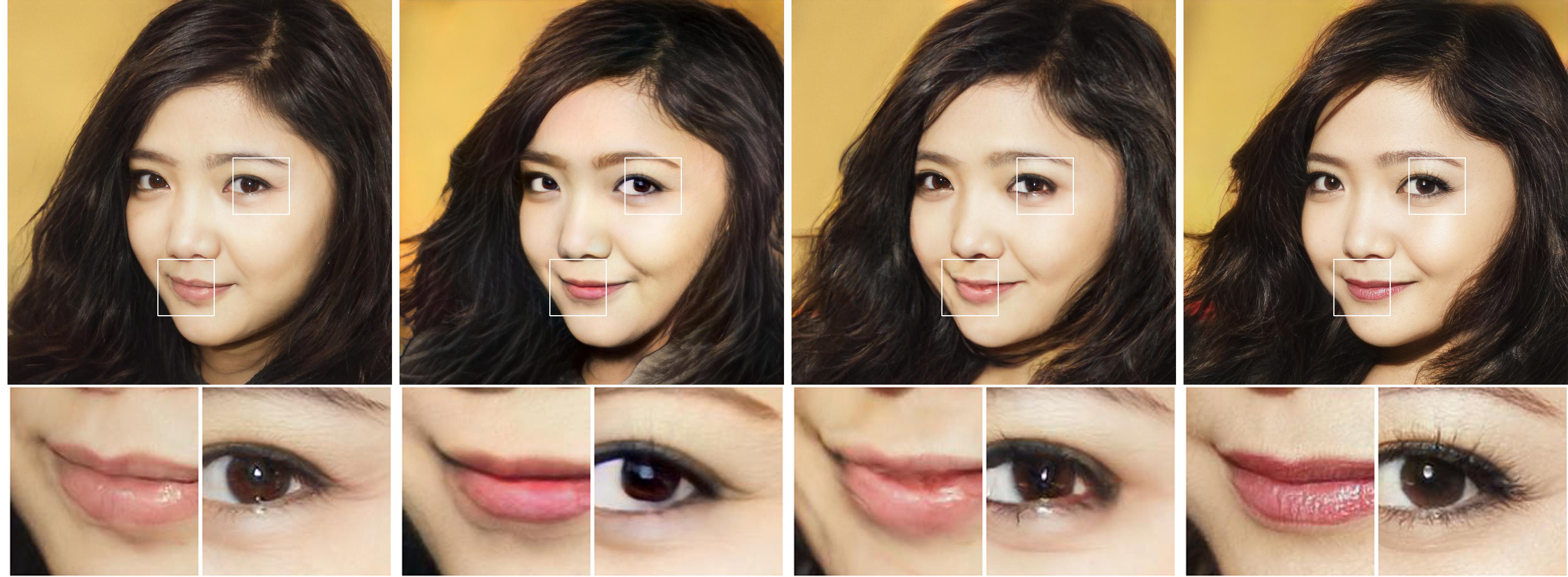}}
\\
\end{tabular}
\caption{\textbf{Inversion on face domain}. We compare our model against state-of-the-art GAN inversion methods~\cite{abdal2020image2stylegan++,zhu2020domain,richardson2020encoding,tov2021designing,alaluf2021restyle,wang2021high} for the inversion of StyleGAN2 pre-trained on face domain. Reconstructions using our framework are visually more faithful and zoom-in patches show that they exhibit much more details and sharpness.
}
\label{inversion_face_4}
\end{figure*}

\begin{figure*}[t]
\centering
\small
\setlength{\tabcolsep}{0pt}
\renewcommand{\arraystretch}{1.0}
\begin{tabular}{P{0.25\linewidth}P{0.25\linewidth}P{0.25\linewidth}P{0.25\linewidth}}
\centering
Source & Optimization~\cite{abdal2020image2stylegan++} & In-domain~\cite{zhu2020domain}& pSp~\cite{richardson2020encoding} 
\\
\multicolumn{4}{c}{\includegraphics[width=\linewidth]{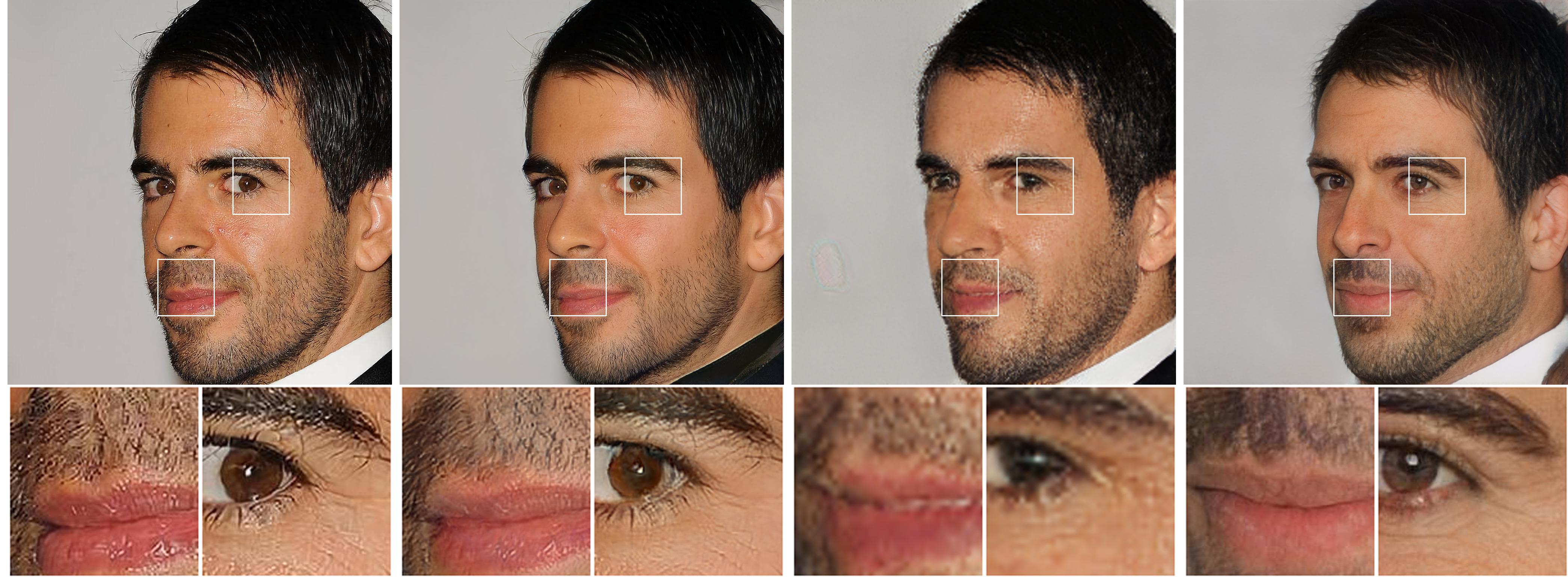}}
\\
e4e~\cite{tov2021designing} & restyle-pSp~\cite{alaluf2021restyle} & HFGI~\cite{wang2021high} & Ours
\\
\multicolumn{4}{c}{\includegraphics[width=\linewidth]{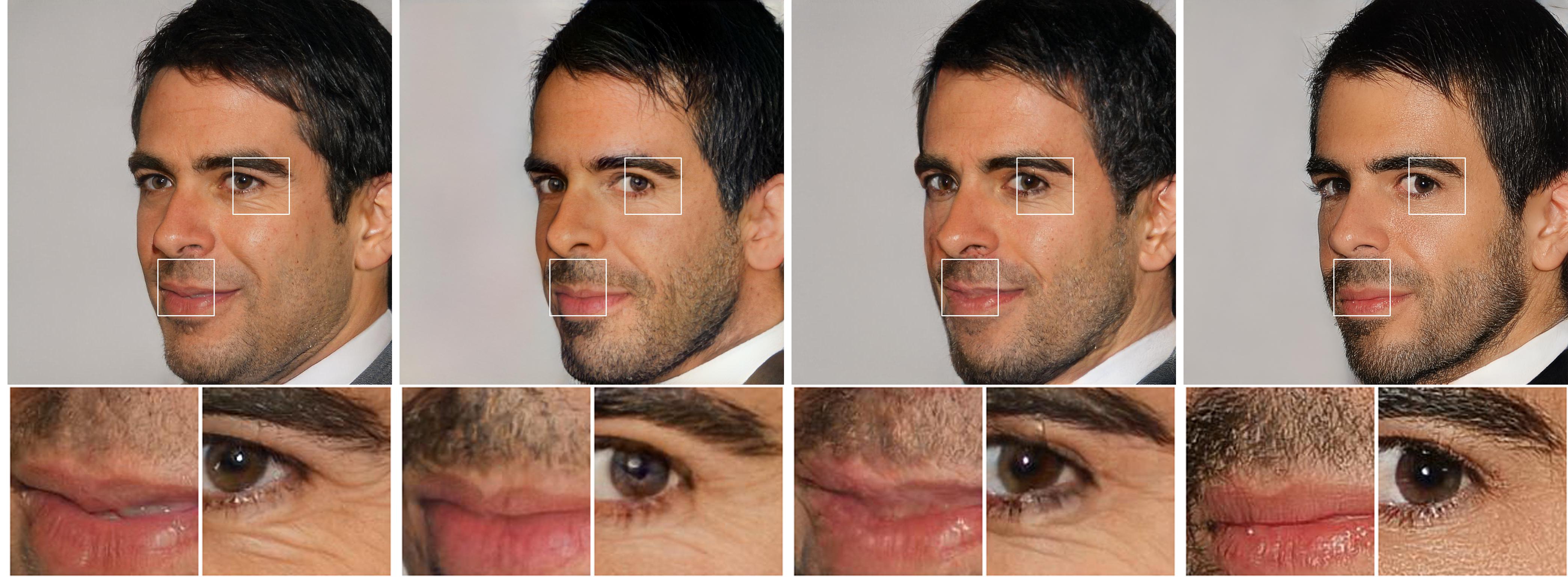}}
\\
\end{tabular}
\caption{\textbf{Inversion on face domain}. We compare our model against state-of-the-art GAN inversion methods~\cite{abdal2020image2stylegan++,zhu2020domain,richardson2020encoding,tov2021designing,alaluf2021restyle,wang2021high} for the inversion of StyleGAN2 pre-trained on face domain. Reconstructions using our framework are visually more faithful and zoom-in patches show that they exhibit much more details and sharpness.
}
\label{inversion_face_5}
\end{figure*}

We show additional visual results for the inversion of StyleGAN2 pre-trained on face domain in~\cref{inversion_face,inversion_face_2,inversion_face_3,inversion_face_4,inversion_face_5}. We compare our model against an optimization based method~\cite{abdal2020image2stylegan++}, state-of-the-art encoder based methods~\cite{richardson2020encoding,tov2021designing,alaluf2021restyle,wang2021high} and a hybrid method~\cite{zhu2020domain}. As can be observed, reconstructions using our framework are visually more faithful and zoom-in patches show that they exhibit much more details and sharpness.

\begin{figure*}[t]
\centering
\small
\setlength{\tabcolsep}{1pt}
\renewcommand{\arraystretch}{0.5}
\begin{tabular}{P{0.24\linewidth}P{0.24\linewidth}P{0.24\linewidth}P{0.24\linewidth}}
\centering
Source & Inversion & Source & Inversion
\\
\multicolumn{2}{c}{\includegraphics[width=0.49\linewidth]{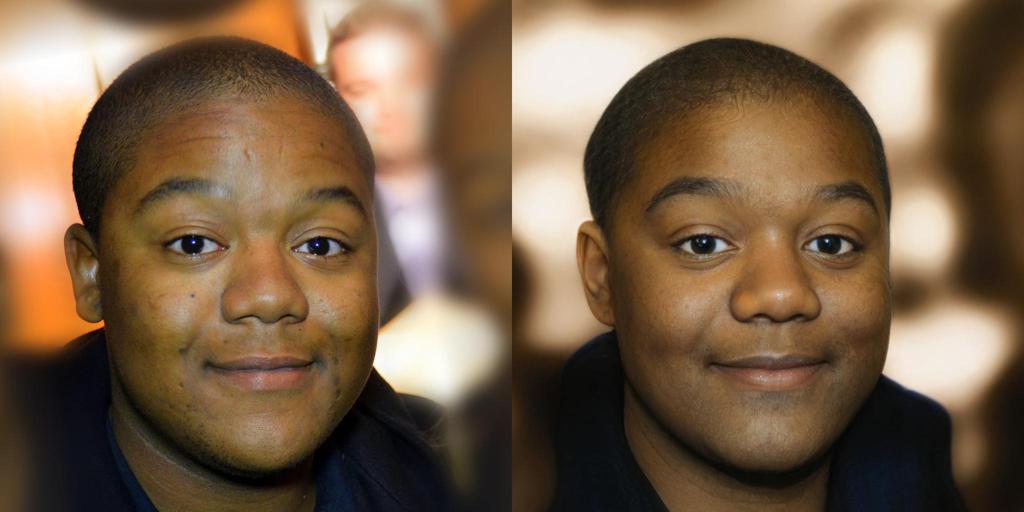}}&
\multicolumn{2}{c}{\includegraphics[width=0.49\linewidth]{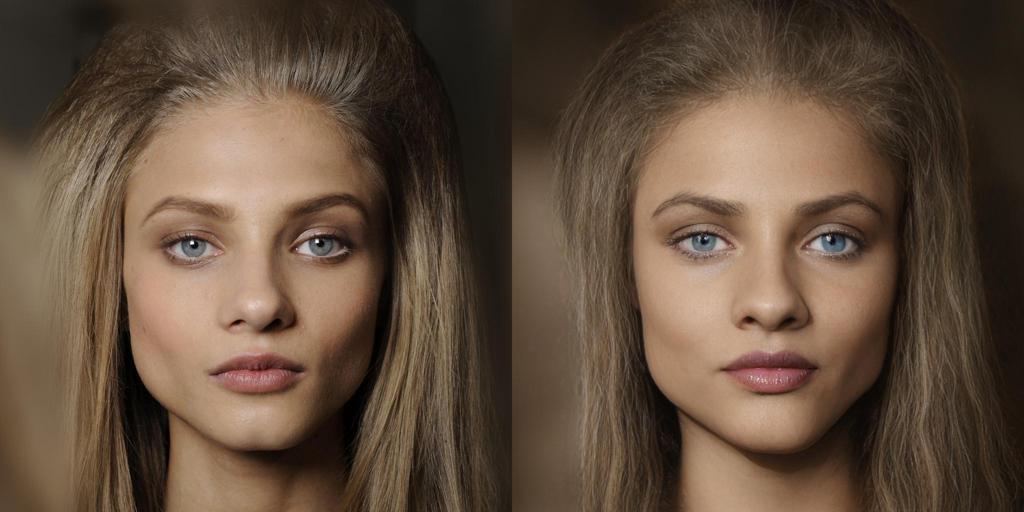}}
\\
\multicolumn{2}{c}{\includegraphics[width=0.49\linewidth]{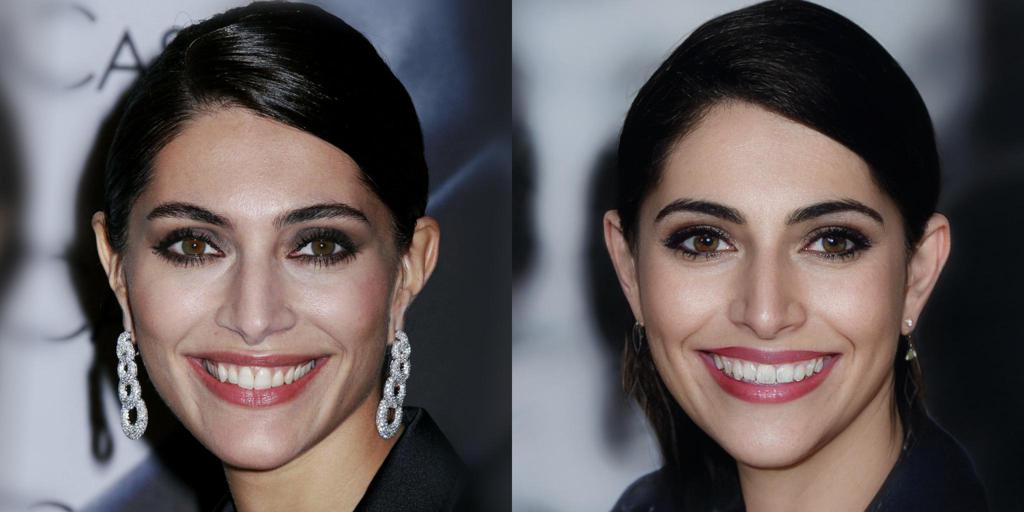}}&
\multicolumn{2}{c}{\includegraphics[width=0.49\linewidth]{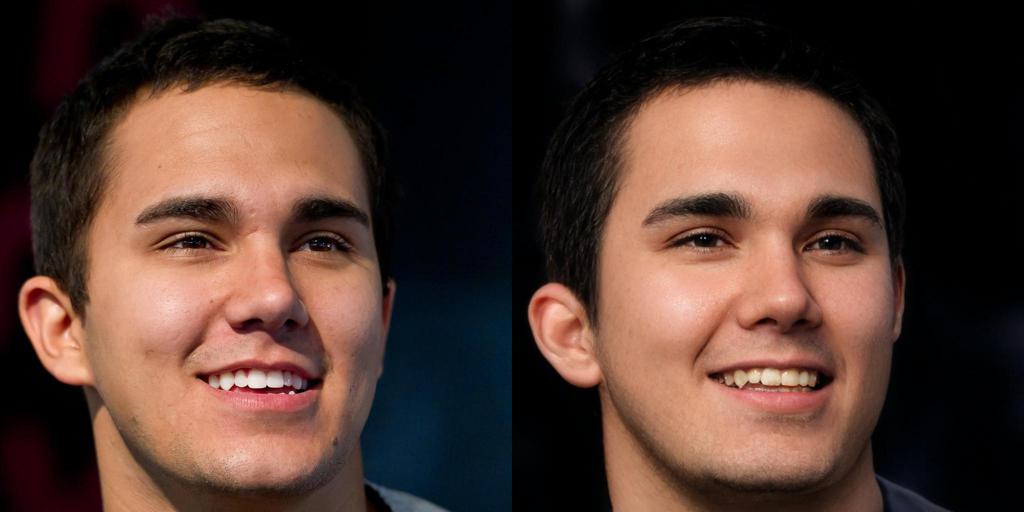}}
\\
\multicolumn{2}{c}{\includegraphics[width=0.49\linewidth]{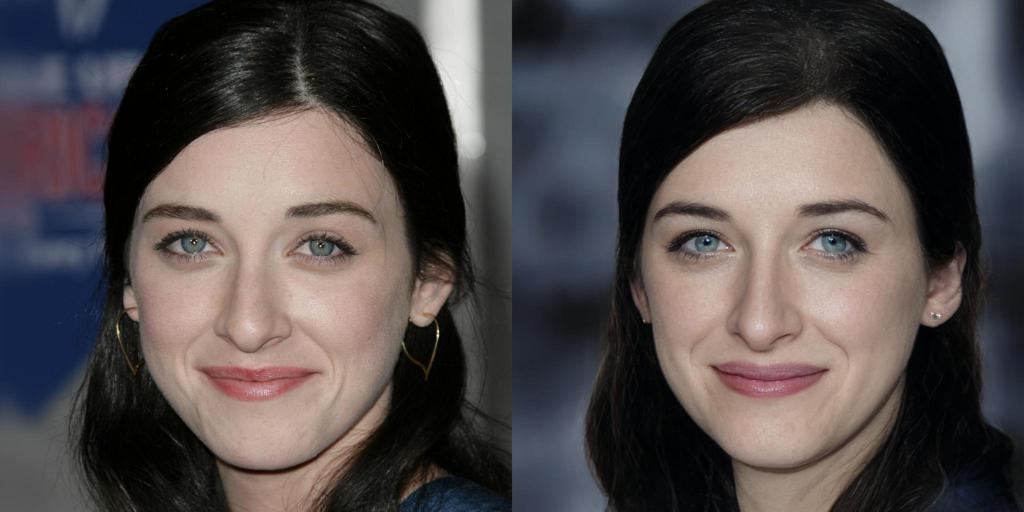}}&
\multicolumn{2}{c}{\includegraphics[width=0.49\linewidth]{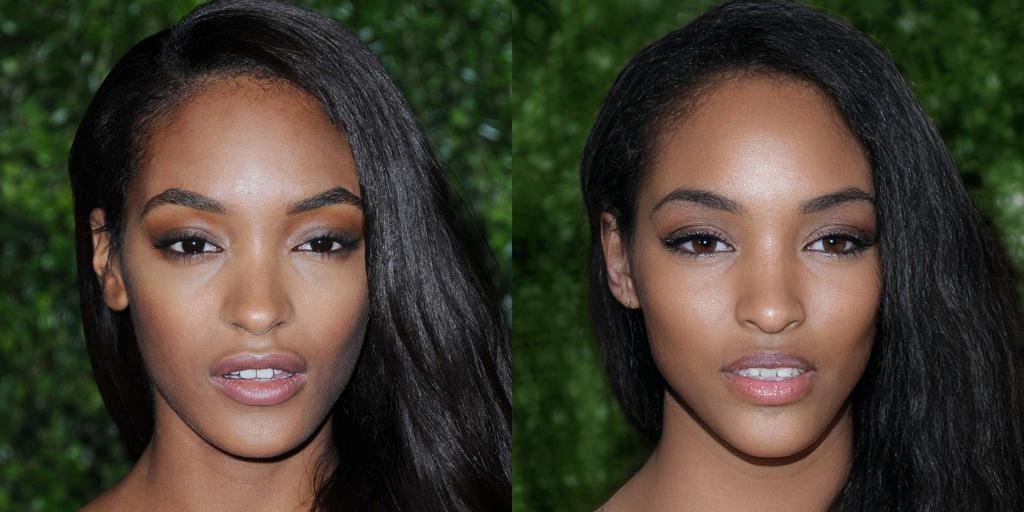}}
\\
\multicolumn{2}{c}{\includegraphics[width=0.49\linewidth]{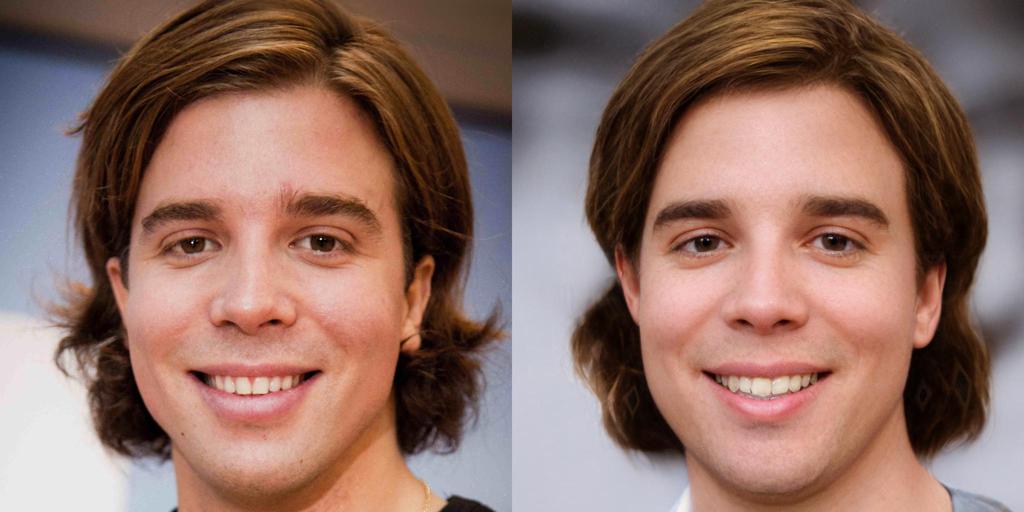}}&
\multicolumn{2}{c}{\includegraphics[width=0.49\linewidth]{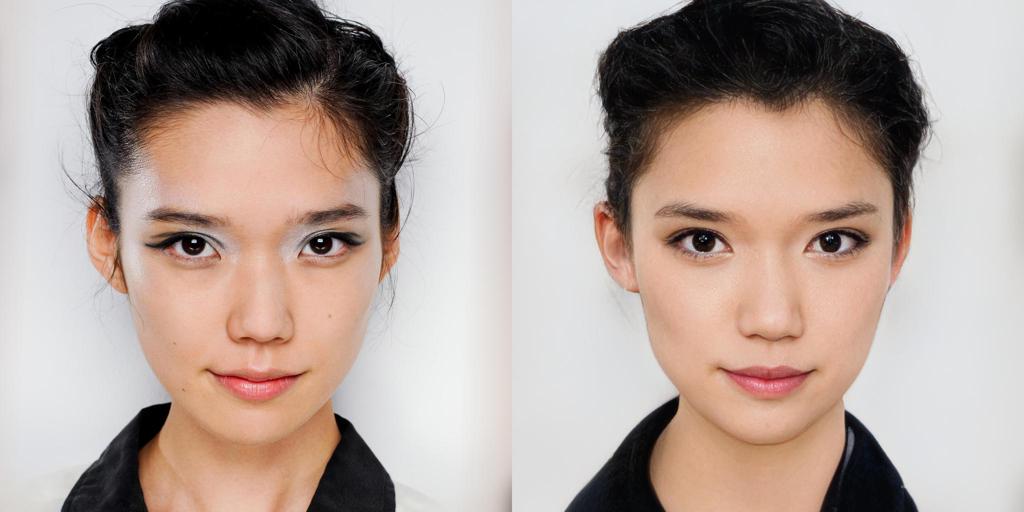}}
\\
\end{tabular} 
\caption{\textbf{Inversion of Alias-free GAN.} 
We show preliminary inversion results of the 3rd generation of StyleGAN - recent released (one month ago) Alias-free GAN \cite{karras2021alias} pretrained on face domain. Compared with StyleGAN2, the architecture of Alias-free GAN has several important changes. Despite the architectural changes, our proposed encoder still yields satisfying inversion results.  
}
\label{inversion_alias}
\end{figure*}

\noindent\textbf{Alias-free GAN}\quad
We show preliminary inversion results of the third generation on StyleGAN - recently released (one month ago) Alias-free GAN~\cite{karras2021alias} pretrained on FFHQ~\cite{karras2019style}. Compared with StyleGAN2, the architecture of Alias-free GAN has several important changes. First, the input tensor passed into the generator is no longer constant, but synthesized from the latent code. The spatial size of the input tensor is increased from $4 \times 4$ to $36 \times 36$. Additionally, the noise inputs are discarded. As shown in Figure~\ref{inversion_alias}, despite the architectural changes, our proposed encoder still yields satisfying inversion results.

\section{Editing}

\noindent\textbf{Latent Space Editing}\quad 
We show additional facial attribute editing results in Figure~\ref{attr_edit}. The latent editing directions are computed using InterFaceGAN~\cite{shen2019interpreting}, except the last attribute `pose', computed with SeFa~\cite{shen2021closed}. 

\begin{figure*}[t]
\centering
\small
\setlength{\tabcolsep}{0pt}
\renewcommand{\arraystretch}{0}
\begin{tabular}{P{0.2\linewidth}P{0.2\linewidth}P{0.2\linewidth}P{0.2\linewidth}P{0.2\linewidth}}
\centering
Source & Inversion & Edit Direction 1 & Edit Direction 2 & Edit Direction 3
\vspace{1pt}
\\
\multicolumn{5}{c}{\includegraphics[width=0.99\linewidth]{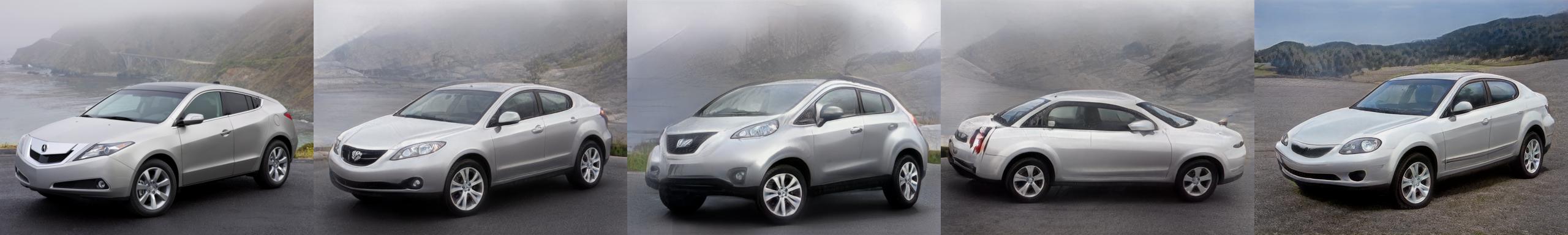}}
\\
\multicolumn{5}{c}{\includegraphics[width=0.99\linewidth]{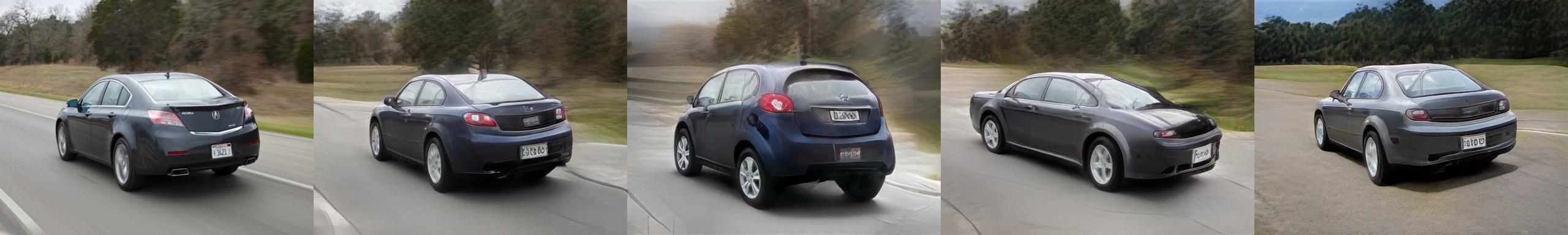}}
\\
\multicolumn{5}{c}{\includegraphics[width=0.99\linewidth]{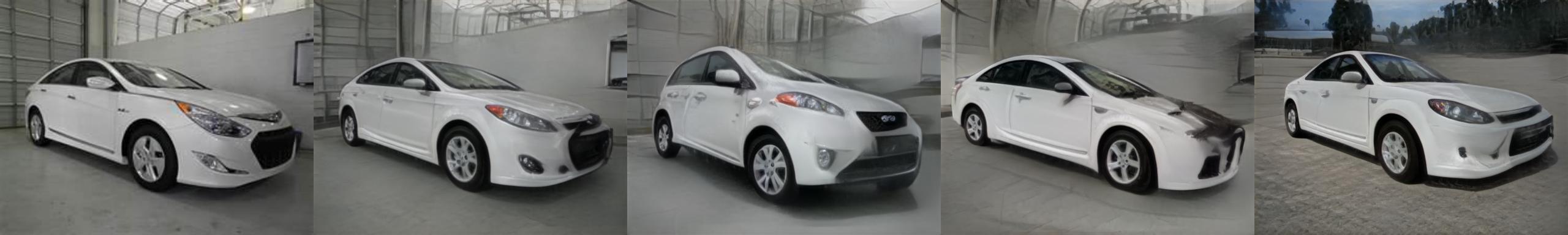}}
\\
\multicolumn{5}{c}{\includegraphics[width=0.99\linewidth]{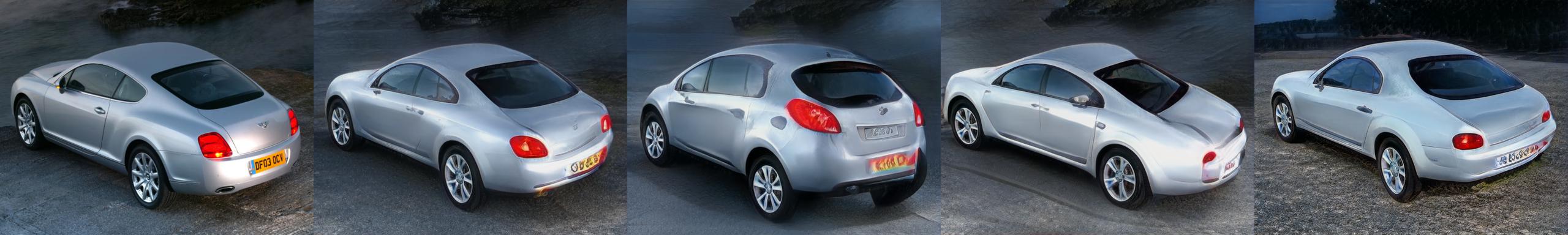}}
\\
\multicolumn{5}{c}{\includegraphics[width=0.99\linewidth]{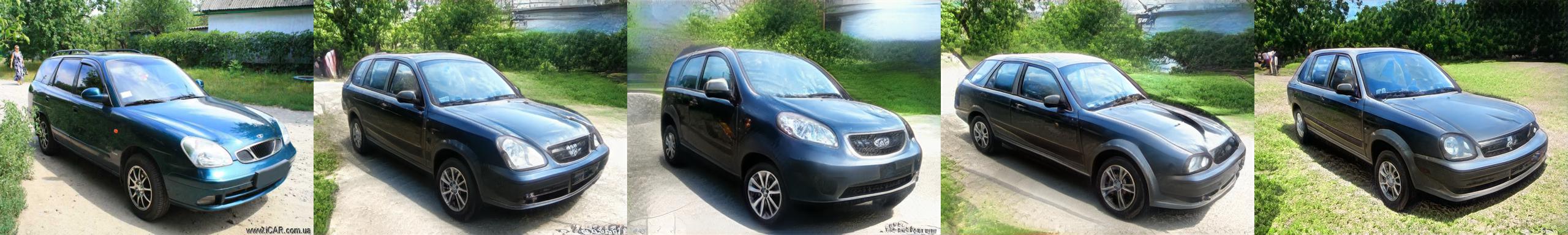}}
\\
\multicolumn{5}{c}{\includegraphics[width=0.99\linewidth]{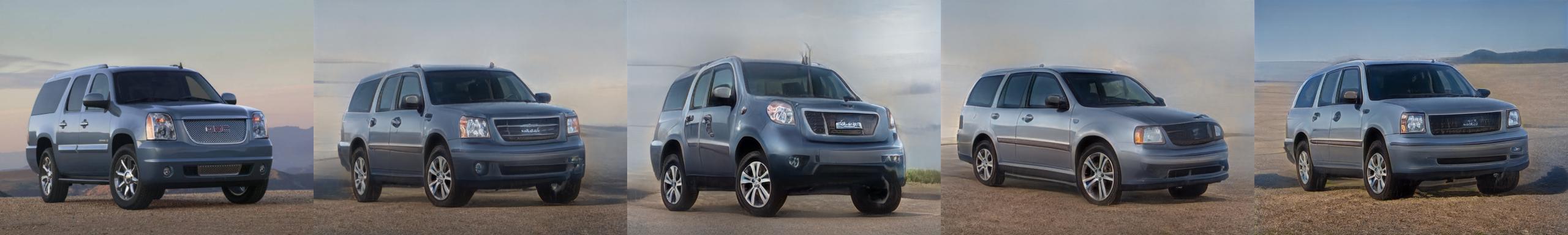}}
\\
\multicolumn{5}{c}{\includegraphics[width=0.99\linewidth]{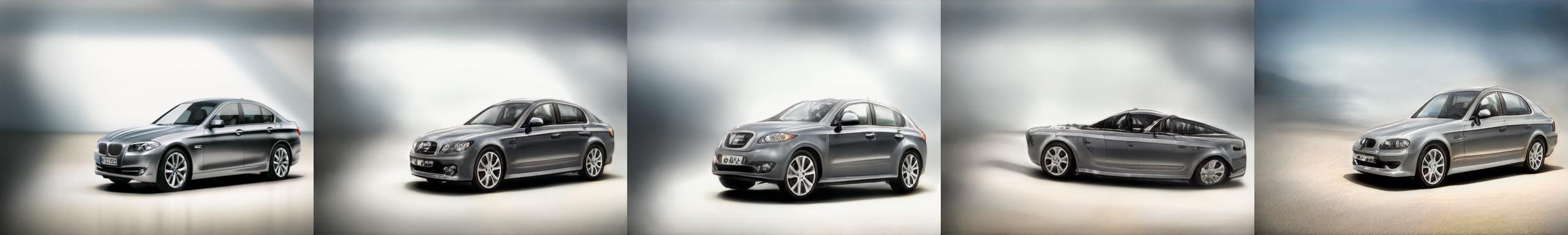}}
\\
\end{tabular}
\caption{\textbf{Editing on car domain}. 
We show latent space editing results on car domain. We compute the latent editing directions with SeFa~\cite{shen2021closed}.
The first column is the source image, second column is our inversion result, the third to last column correspond to the semantic directions found with SeFa~\cite{shen2021closed}. Our model yields satisfying editing results on car domain.
}
\label{edit_car}
\end{figure*}

\begin{figure*}[t]
\centering
\small
\setlength{\tabcolsep}{0pt}
\renewcommand{\arraystretch}{0}
\begin{tabular}{P{0.2\linewidth}P{0.2\linewidth}P{0.2\linewidth}P{0.2\linewidth}P{0.2\linewidth}}
\centering
Source & Inversion & Edit Direction 1 & Edit Direction 2 & Edit Direction 3
\vspace{1pt}
\\
\multicolumn{5}{c}{\includegraphics[width=0.99\linewidth]{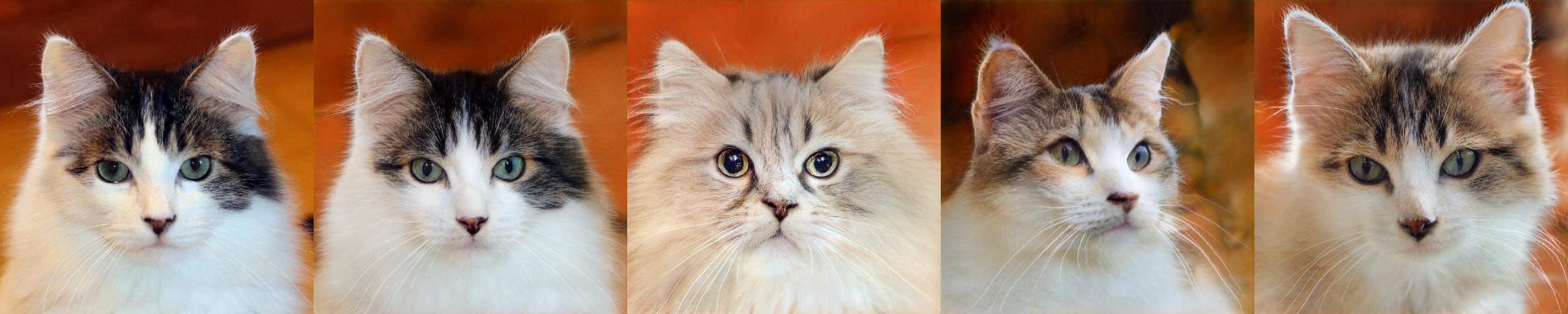}}
\\
\multicolumn{5}{c}{\includegraphics[width=0.99\linewidth]{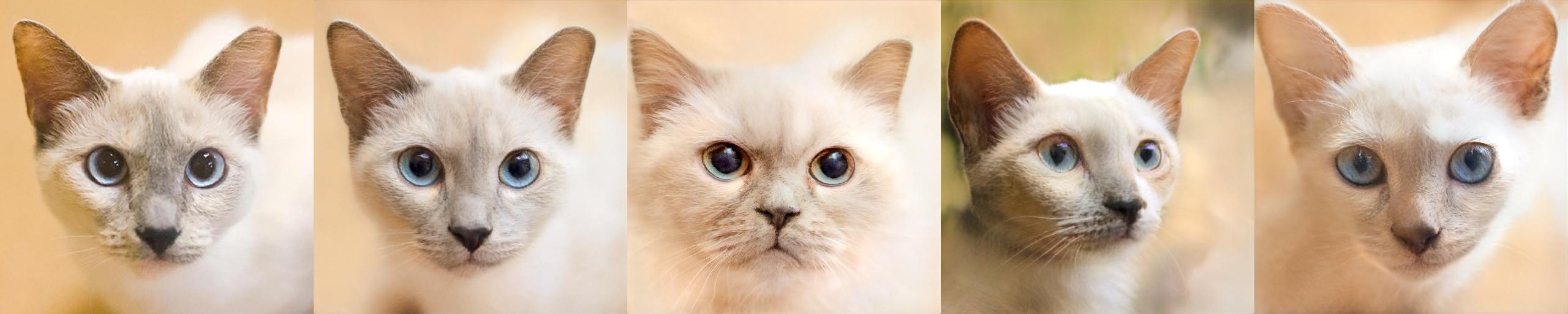}}
\\
\multicolumn{5}{c}{\includegraphics[width=0.99\linewidth]{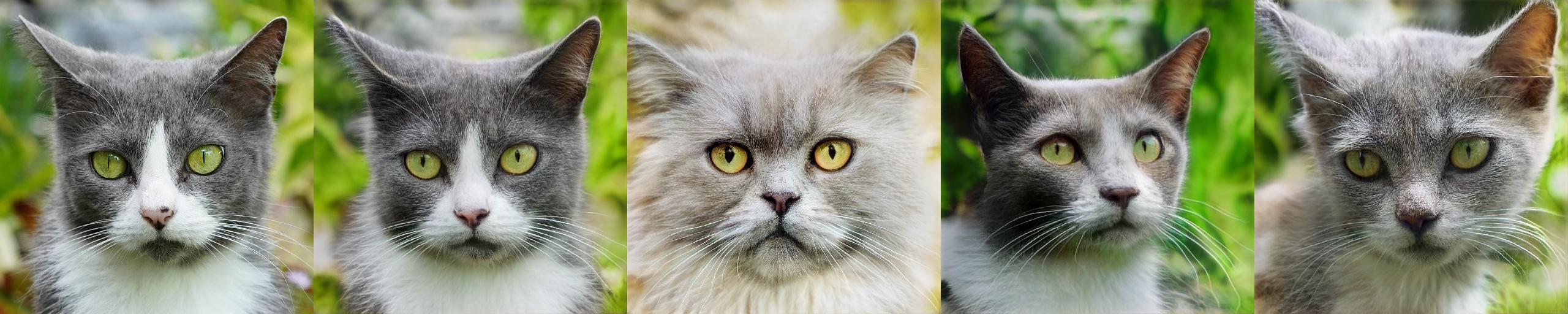}}
\\
\multicolumn{5}{c}{\includegraphics[width=0.99\linewidth]{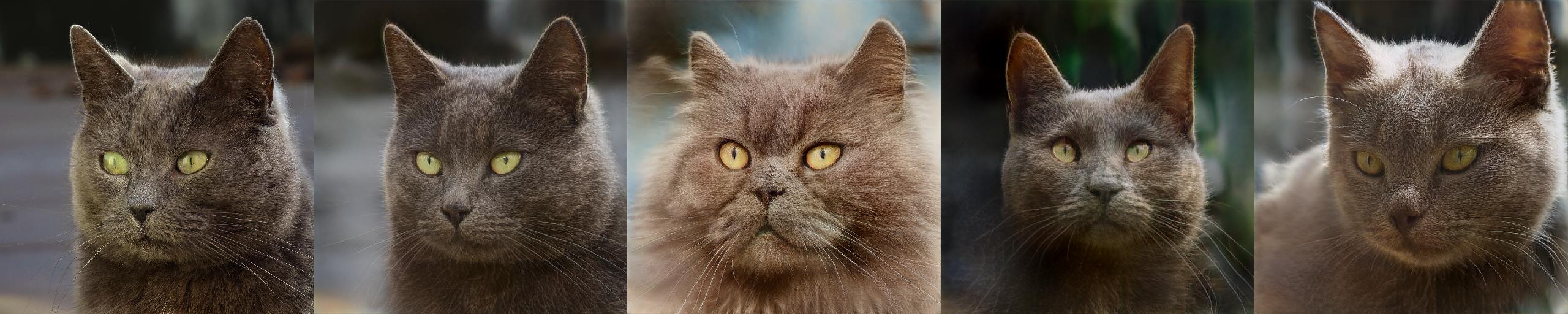}}
\\
\multicolumn{5}{c}{\includegraphics[width=0.99\linewidth]{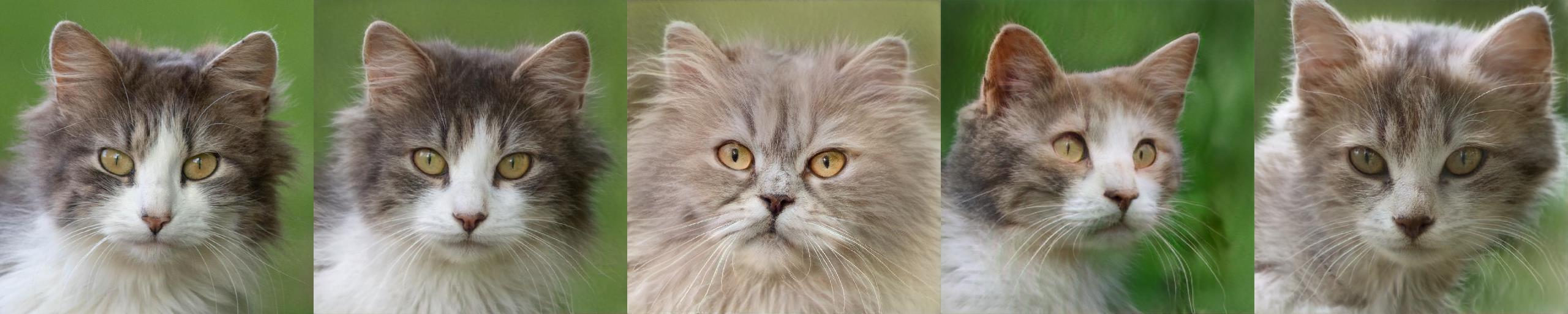}}
\\
\end{tabular}
\caption{\textbf{Editing on cat domain}. 
We show latent space editing results on cat domain. We compute the latent editing directions with SeFa~\cite{shen2021closed}.
The first column is the source image, second column is our inversion result, the third to last column correspond to the semantic directions found with SeFa~\cite{shen2021closed}. Our model yields satisfying editing results on cat domain.
}
\label{edit_cat}
\end{figure*}

\begin{figure*}[t]
\centering
\small
\setlength{\tabcolsep}{0pt}
\renewcommand{\arraystretch}{0}
\begin{tabular}{P{0.2\linewidth}P{0.2\linewidth}P{0.2\linewidth}P{0.2\linewidth}P{0.2\linewidth}}
\centering
Source & Inversion & Edit Direction 1 & Edit Direction 2 & Edit Direction 3
\vspace{1pt}
\\
\multicolumn{5}{c}{\includegraphics[width=0.99\linewidth]{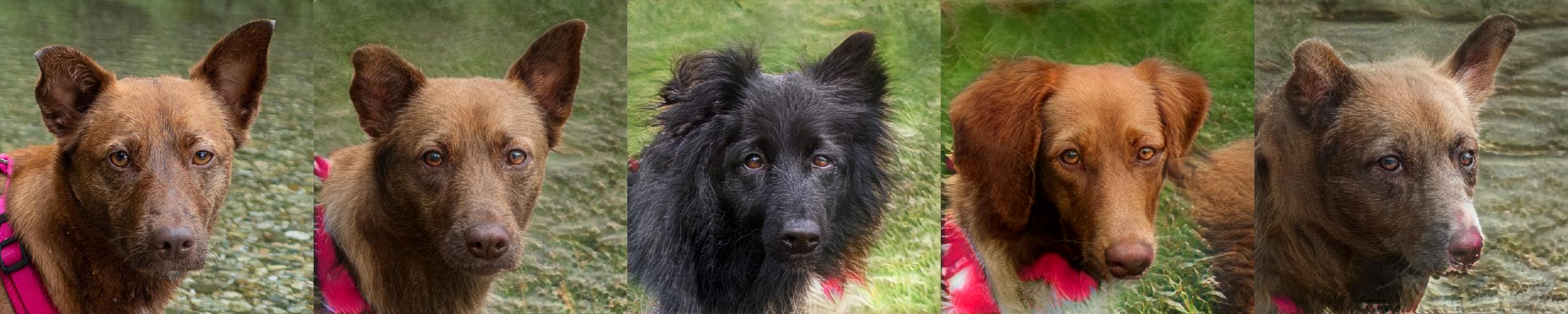}}
\\
\multicolumn{5}{c}{\includegraphics[width=0.99\linewidth]{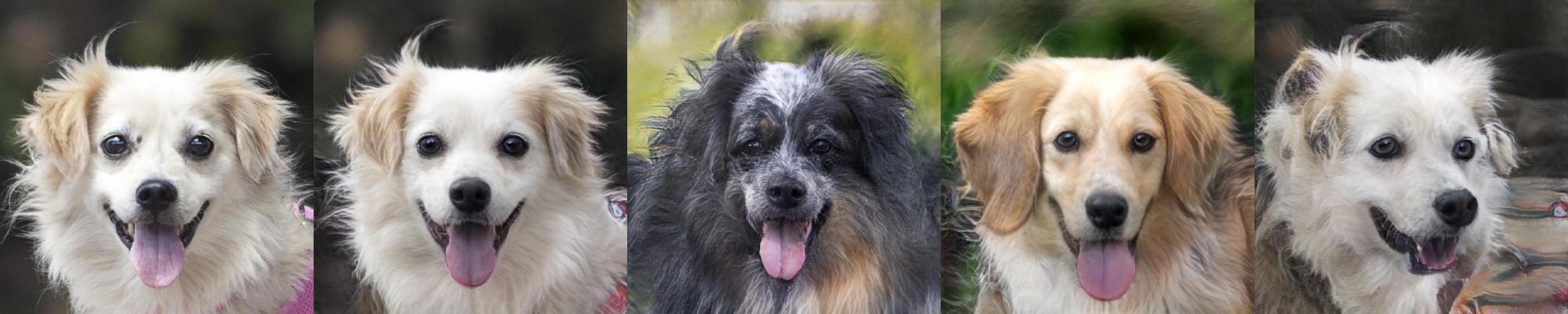}}
\\
\multicolumn{5}{c}{\includegraphics[width=0.99\linewidth]{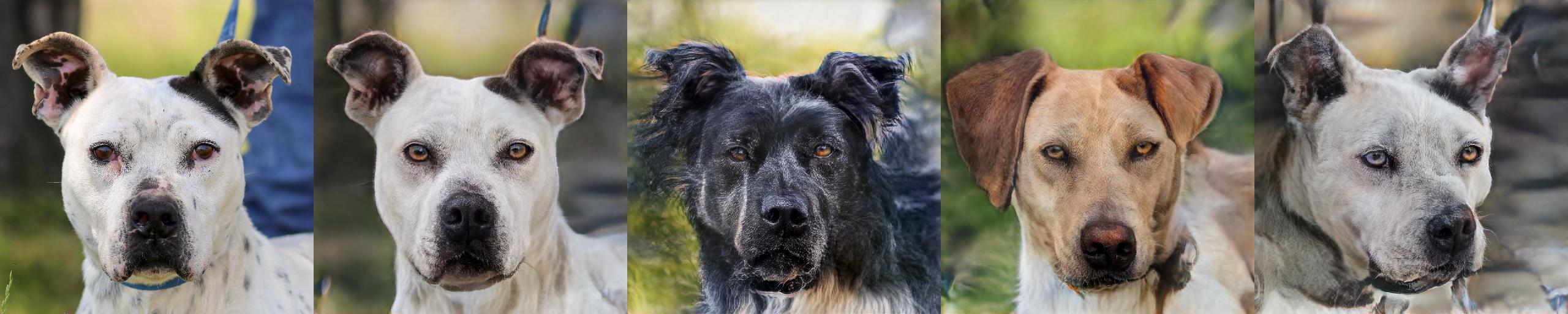}}
\\
\multicolumn{5}{c}{\includegraphics[width=0.99\linewidth]{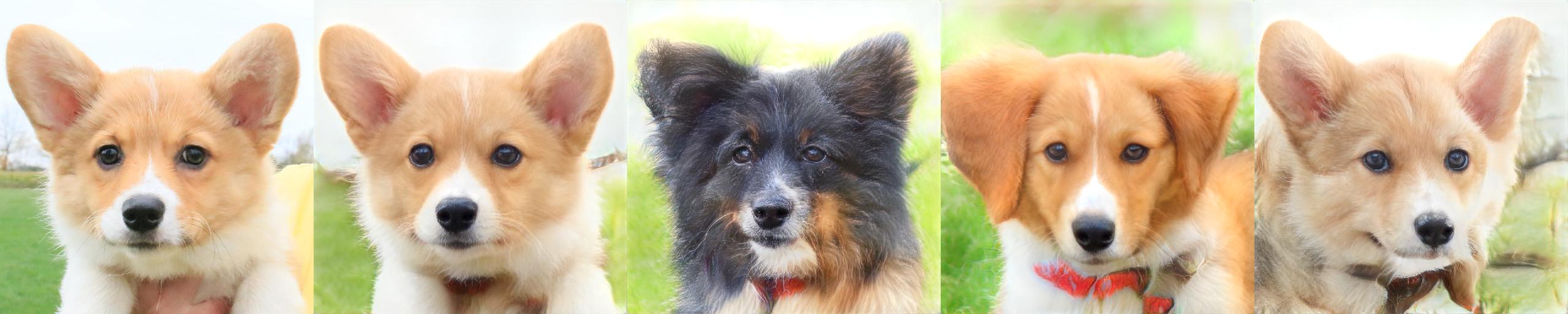}}
\\
\multicolumn{5}{c}{\includegraphics[width=0.99\linewidth]{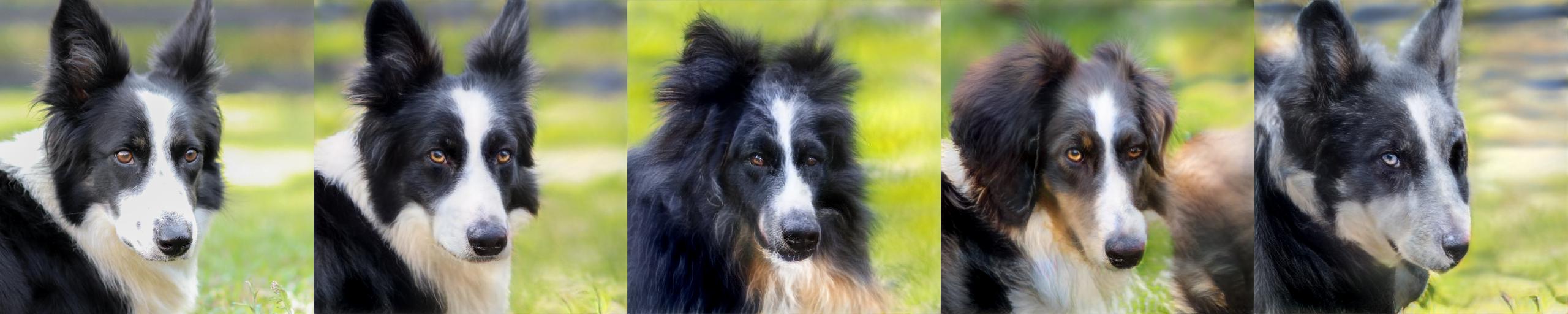}}
\\
\end{tabular}
\caption{\textbf{Editing on dog domain}. 
We show latent space editing results on dog domain. We compute the latent editing directions with SeFa~\cite{shen2021closed}.
The first column is the source image, second column is our inversion result, the third to last column correspond to the semantic directions found with SeFa~\cite{shen2021closed}. Our model yields satisfying editing results on dog domain.
}
\label{edit_dog}
\end{figure*}

\noindent\textbf{Editing on Other Domains}\quad
In the main paper, we have presented the inversion results for StyleGAN2 pretrained on the car domain and StyleGAN2-Ada pretrained on cat and dog domain. Here we present additional editing results.
Figure~\ref{edit_car} shows editing results on car domain. 
Figure~\ref{edit_cat} and Figure~\ref{edit_dog} show the editing results on cat and dog domain, respectively. All the latent editing directions are computed with SeFa. Our model yields satisfying editing results on other domains.

\begin{figure*}[t]
\centering
\small
\setlength{\tabcolsep}{0pt}
\renewcommand{\arraystretch}{1.0}
\begin{tabular}{P{0.24\linewidth}P{0.24\linewidth}P{0.24\linewidth}P{0.24\linewidth}}
\centering
$\ve{G}(\ve{w}_A,\ve{F}_A)$ & $\ve{G}(\ve{w}_B,\ve{F}_A)$ & $\ve{G}(\ve{w}_A,\ve{F}_B)$ & $\ve{G}(\ve{w}_B,\ve{F}_B)$
\\
\multicolumn{4}{c}{\includegraphics[width=0.95\linewidth]{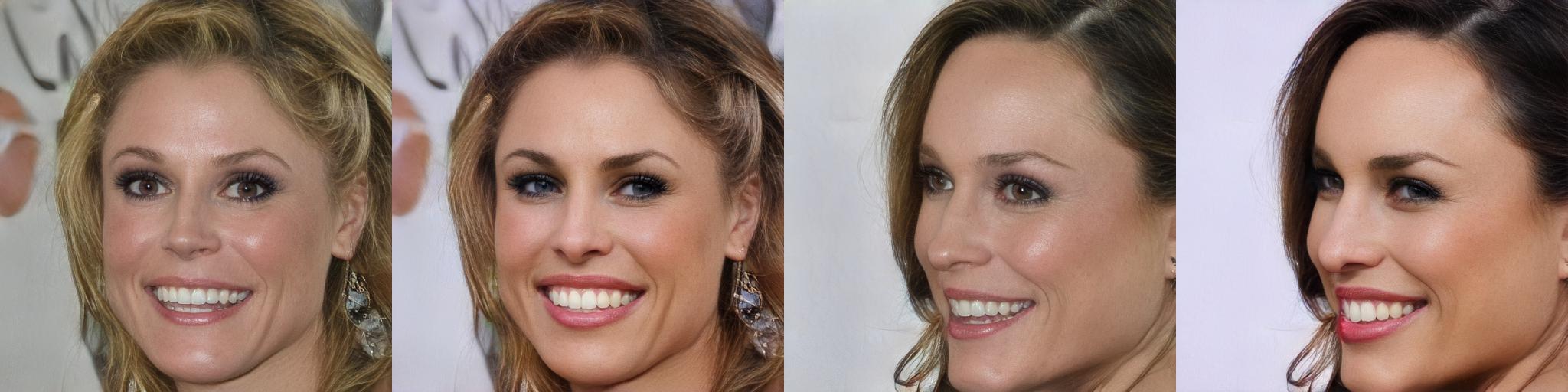}}
\\
$\ve{G}(\ve{w}_A,\ve{F}_A)$ & $\ve{G}(\ve{w}_B,\ve{F}_A)$ & $\ve{G}(\ve{w}_A,\ve{F}_B)$ & $\ve{G}(\ve{w}_B,\ve{F}_B)$
\\
\multicolumn{4}{c}{\includegraphics[width=0.95\linewidth]{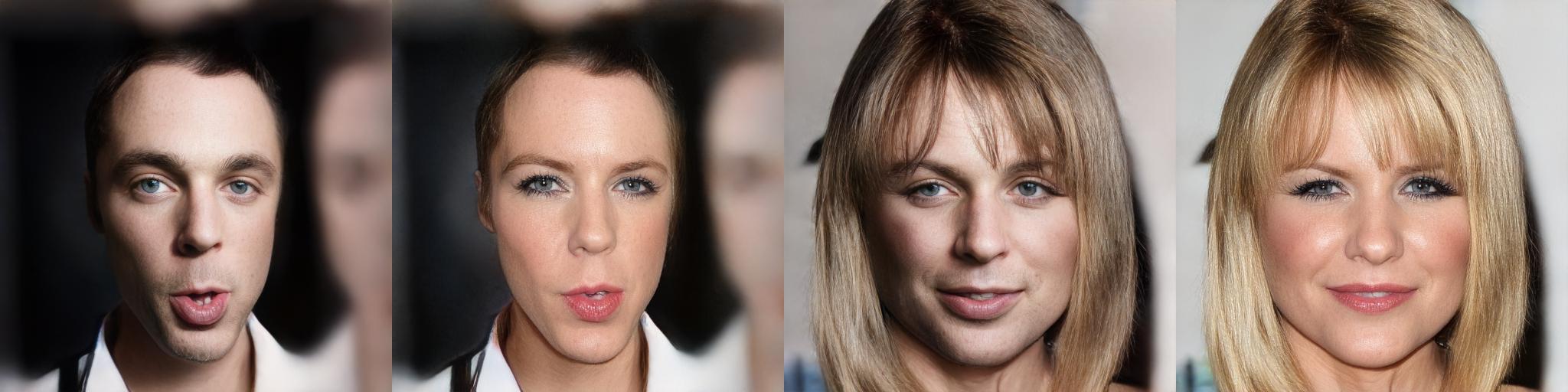}}
\\
$\ve{G}(\ve{w}_A,\ve{F}_A)$ & $\ve{G}(\ve{w}_B,\ve{F}_A)$ & $\ve{G}(\ve{w}_A,\ve{F}_B)$ & $\ve{G}(\ve{w}_B,\ve{F}_B)$
\\
\multicolumn{4}{c}{\includegraphics[width=0.95\linewidth]{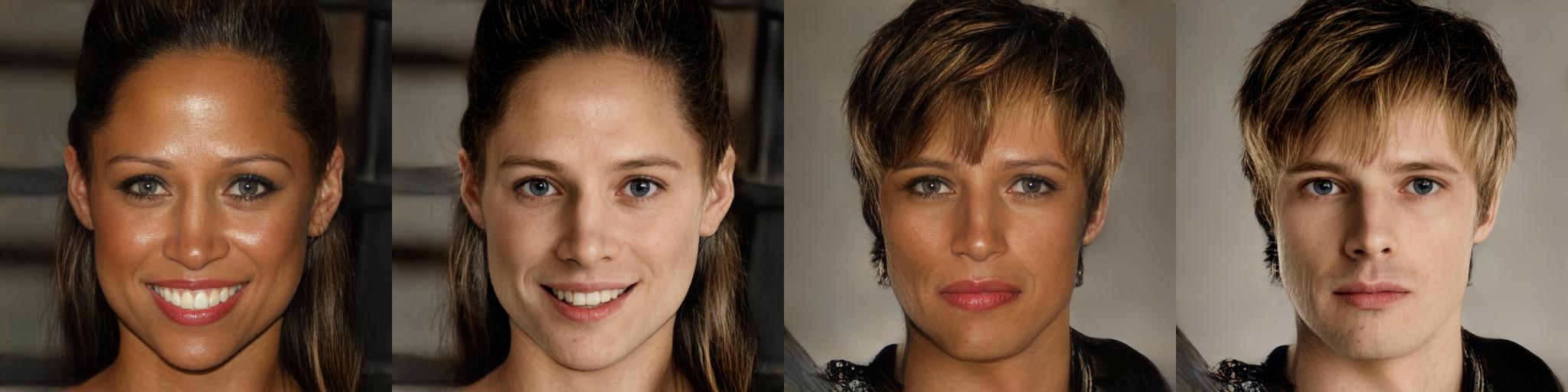}}
\\
\end{tabular} 
\caption{\textbf{Style mixing of feature code and latent code.} 
The first and last column show the inversions of two images $\ve{x}_A$ and $\ve{x}_B$, denoted by $\ve{G}(\ve{w}_A,\ve{F}_A)$ and $\ve{G}(\ve{w}_B,\ve{F}_B)$, respectively. The second column is generated from the feature code of $\ve{x}_A$ and the latent code of $\ve{x}_B$, denoted by $\ve{G}(\ve{w}_B,\ve{F}_A)$, and vice versa for the third column, denoted by $\ve{G}(\ve{w}_A,\ve{F}_B)$. The feature code encodes the geometric structures such as pose and facial shape, whereas the latent code controls the appearance styles like eye color and makeup.
}
\label{style_mix}
\end{figure*}

\noindent\textbf{Style Mixing}\quad 
We show additional style mixing results in Figure~\ref{style_mix}. The style mixing is generated from the latent code of one image with the feature code of another image. These additional results confirm that the geometric structures such as pose and facial shape are encoded by the feature code, while the appearance styles like eye color and makeup are encoded by the latent code.

\section{Video results}

\begin{figure}[t]
\centering
\includegraphics[width=0.99\linewidth]{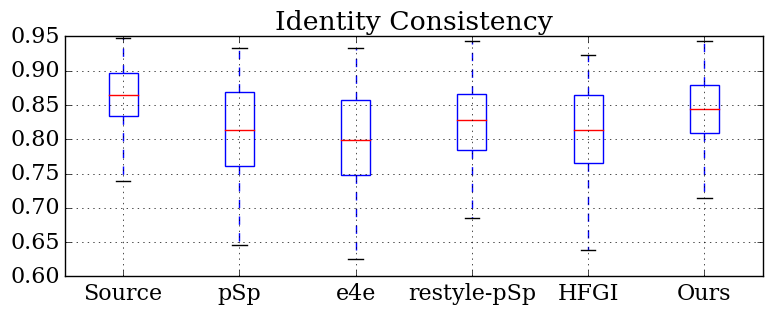}
\caption{\textbf{Identity consistency of video inversion}. For each method, we compute the proposed metric \emph{identity consistency} for each inverted video and plot the results in a box-plot. Our averaged identity consistency is the closest to that of the source videos.
}
\label{id_consistency}
\end{figure}

\noindent\textbf{Inversion Consistency}\quad 
Additionally, to evaluate the consistency of the inversion, we propose a new metric \emph{Identity Consistency}, which refers to the averaged identity similarity between the reconstructed frame $\tilde{\ve{x}}_i$ and frame $\tilde{\ve{x}}_0$ along a video sequence:
\begin{equation}
    IC=\frac{1}N \sum_{i=1}^{N} \langle \ve{R}(\tilde{\ve{x}}_i), \ve{R}(\tilde{\ve{x}}_0) \rangle,
\end{equation}
where \ve{R} is the pre-trained ArcFace\cite{deng2019arcface} network. We compute this metric for our encoder and state-of-the-art methods for video inversion on RAVDESS~\cite{livingstone2018ryerson}. From this dataset we sample randomly 120 videos as evaluation data. For each method, we perform the inversion and compute this metric on each inverted video and present the results with a box-plot. Figure~\ref{id_consistency} shows that the averaged identity consistency of our inversion is the closest to that of the source, which proves the stability of our inversion. 

\section{Ablation study}

\begin{figure*}[t]
\centering
\small
\setlength{\tabcolsep}{0pt}
\renewcommand{\arraystretch}{1.0}
\begin{tabular}{P{0.33\linewidth}P{0.33\linewidth}P{0.33\linewidth}}
\centering
Source 
& (A) w/o $\mathcal{L}_{m\_lpips}$  
& (B) w/o feature input
\\
\multicolumn{3}{c}{\includegraphics[width=\linewidth]{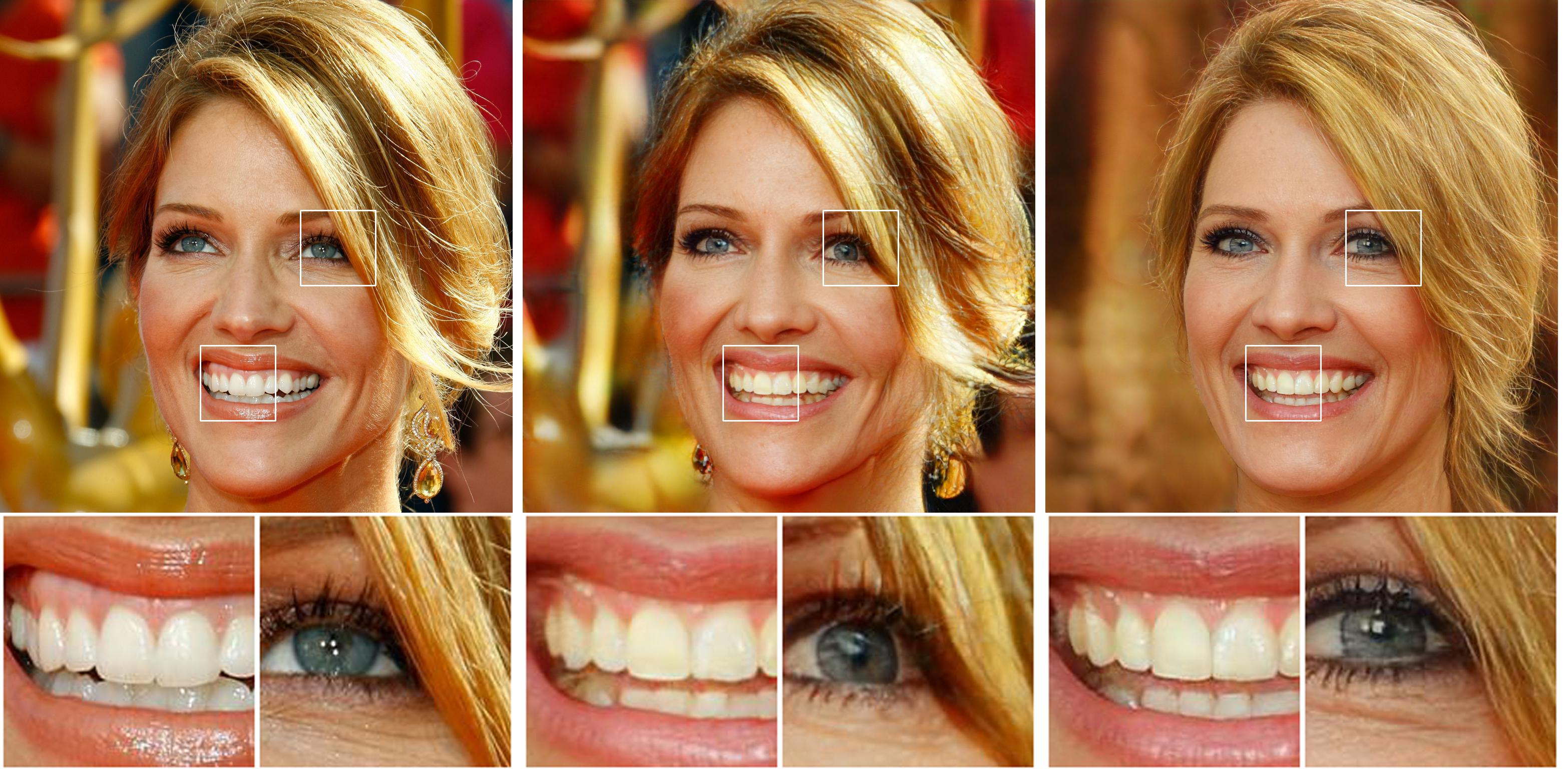}}
\\
& (C) w/o synthetic training data 
& (D) our baseline
\\
&\multicolumn{2}{c}{\includegraphics[width=0.66\linewidth]{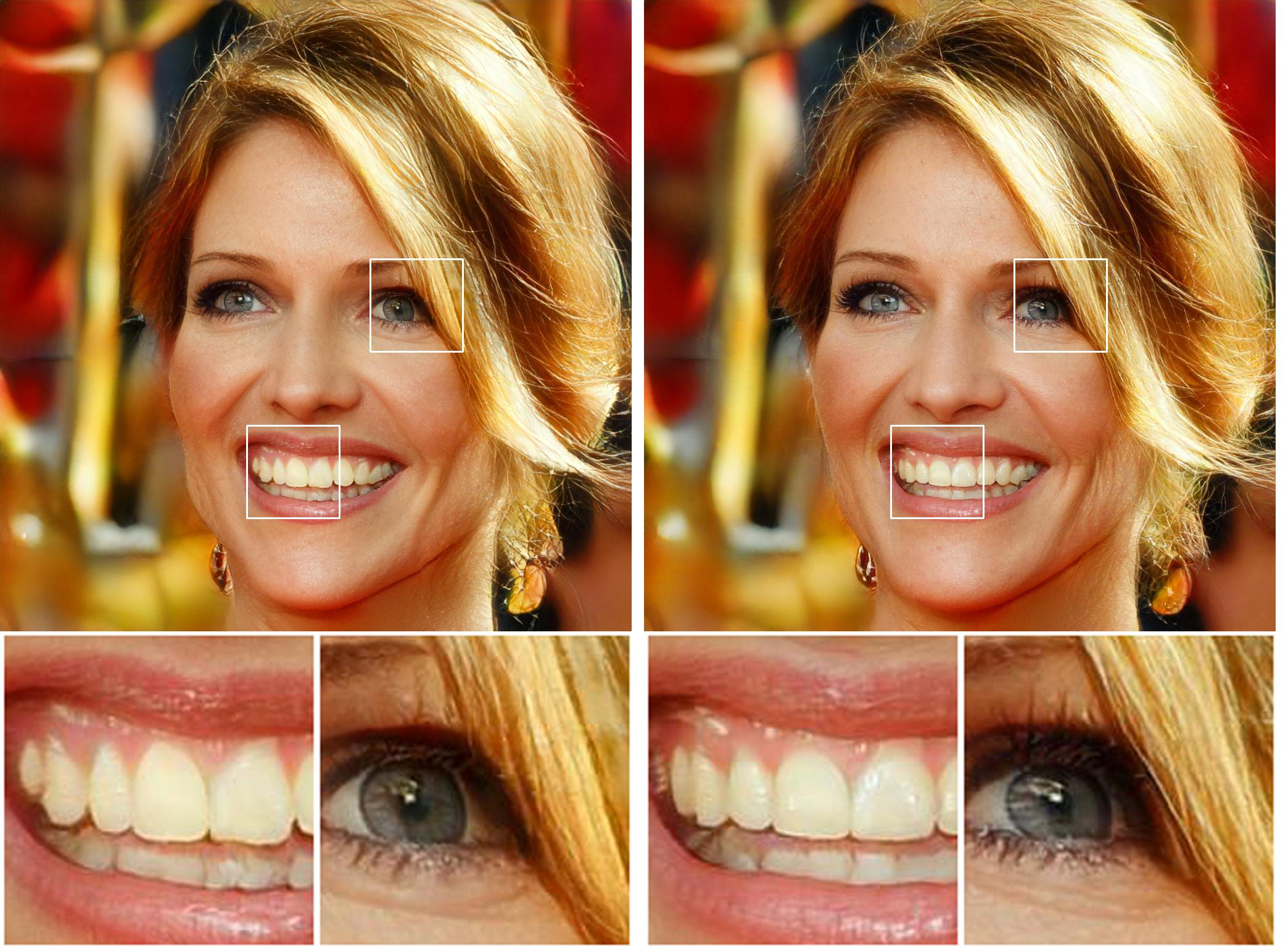}}
\\
\end{tabular}
\caption{\textbf{Visual results of ablation study}. We show the qualitative results of the different ablative configurations. Compared with the inversion result of our baseline, that of configuration (A) is less sharp and reconstructs less well the details. Configuration (B) fails to achieve a plausible reconstruction. The result of configuration (C) is globally plausible, yet less reliable in the details. For instance, the teeth are less photo-realistic compared with our baseline. This experiment confirms the quantitative evaluation in the main paper.
}
\label{ablation_visual}
\end{figure*}

\noindent\textbf{Visual Results}\quad 
In the main paper, we have conducted an ablation study on the experimental setup using quantitative metrics. We have compared the following configurations:
\begin{description}
\setlength\itemsep{0em}
    \item (A) w/o multi-scale setting in the perceptual loss
    \item (B) w/o feature prediction branch
    \item (C) w/o synthetic training data
    \item (D) our baseline
\end{description}
We show additional qualitative results of these ablative configurations in Figure~\ref{ablation_visual}. Compared with our baseline, the inversion of configuration (A) is less sharp and reconstructs less well the details. Configuration (B) fails to achieve a plausible reconstruction. The inversion of configuration (C) is globally plausible, yet less reliable in details. For instance, the teeth are less photo-realistic compared with our baseline. This qualitative comparison confirms the quantitative evaluation in the main paper.

\begin{figure*}[t]
\centering
\small
\setlength{\tabcolsep}{0pt}
\renewcommand{\arraystretch}{0.5}
\begin{tabular}{P{0.04\linewidth}P{0.19\linewidth}P{0.19\linewidth}P{0.19\linewidth}P{0.19\linewidth}P{0.19\linewidth}}
\centering
&Source & $K=4$ & $K=5$ (ours) & $K=6$ & $K=7$
\\
\rotatebox{90}{
\begin{tabular}{*{3}{>{\centering\arraybackslash}m{92pt}}}
Style Mixing with reference B
& Style Mixing with reference A
& Inversion \\
\end{tabular}
}
&
\includegraphics[width=\linewidth]{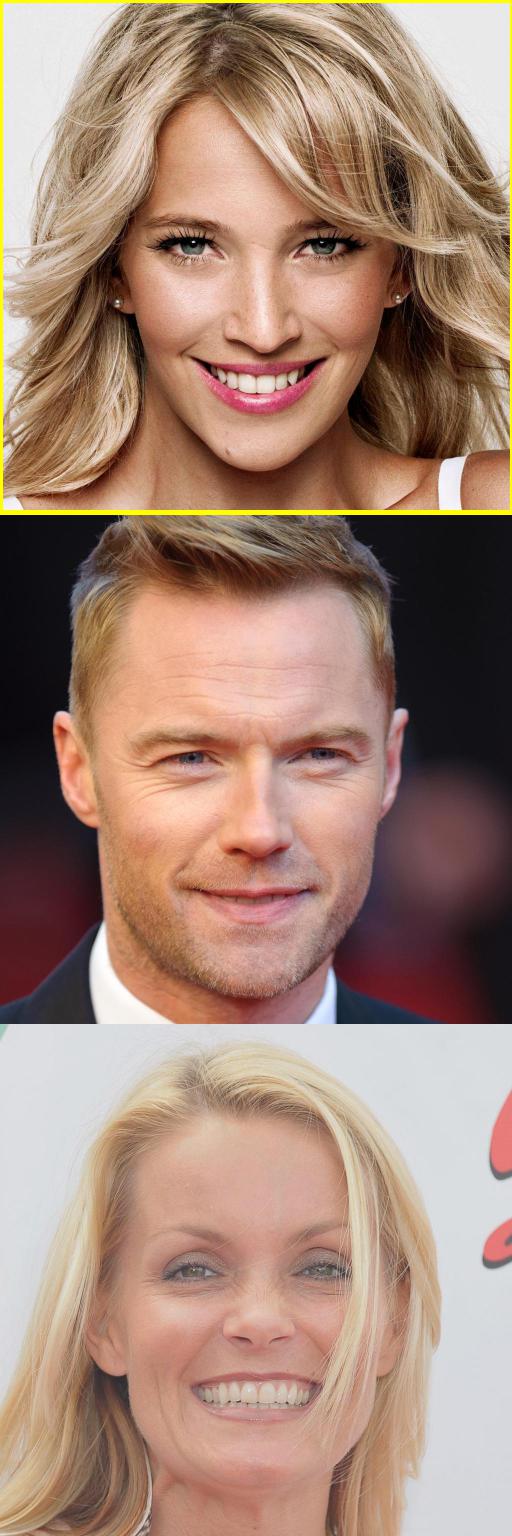}
&
\includegraphics[width=\linewidth]{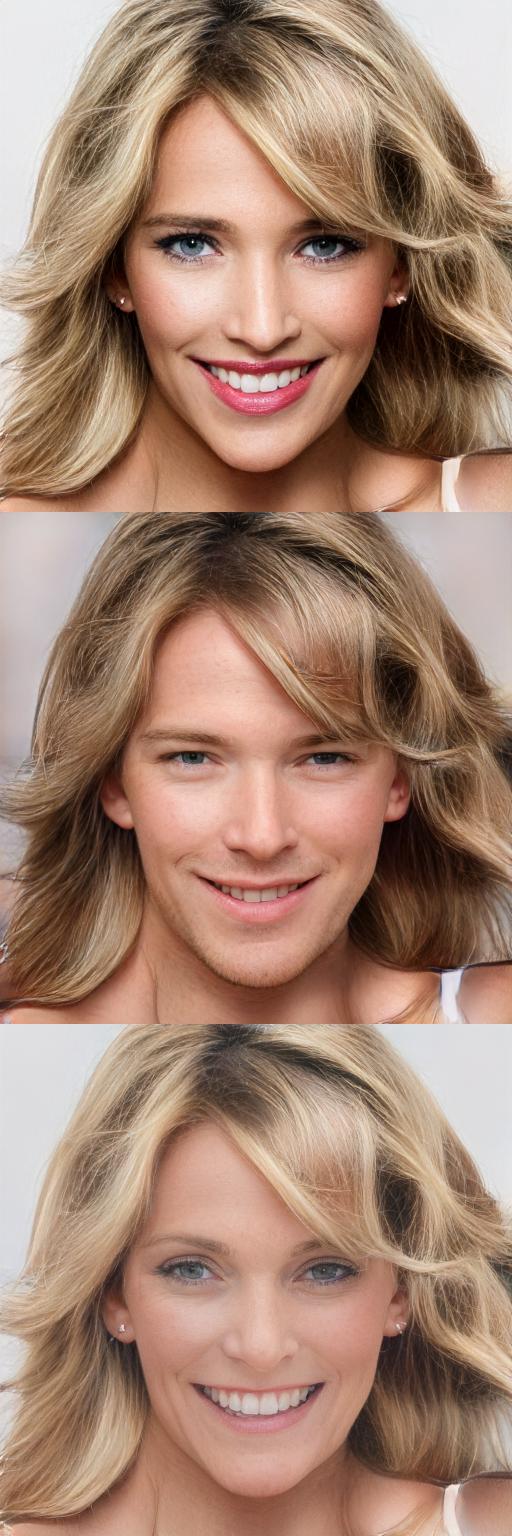}
&
\includegraphics[width=\linewidth]{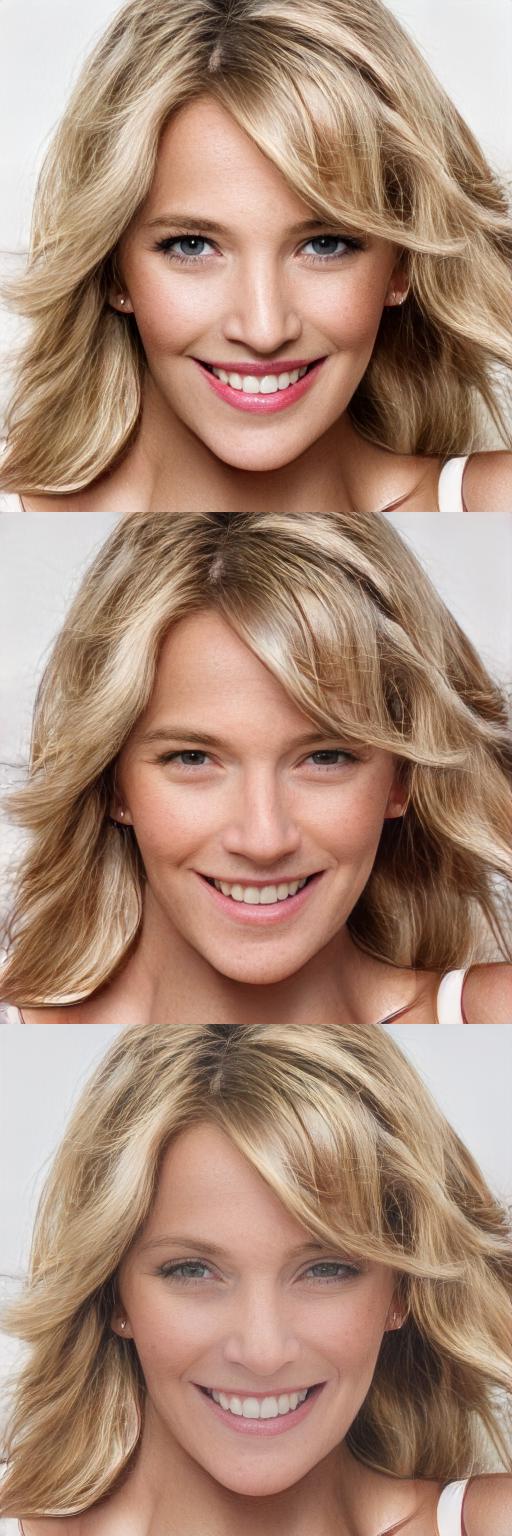}
&
\includegraphics[width=\linewidth]{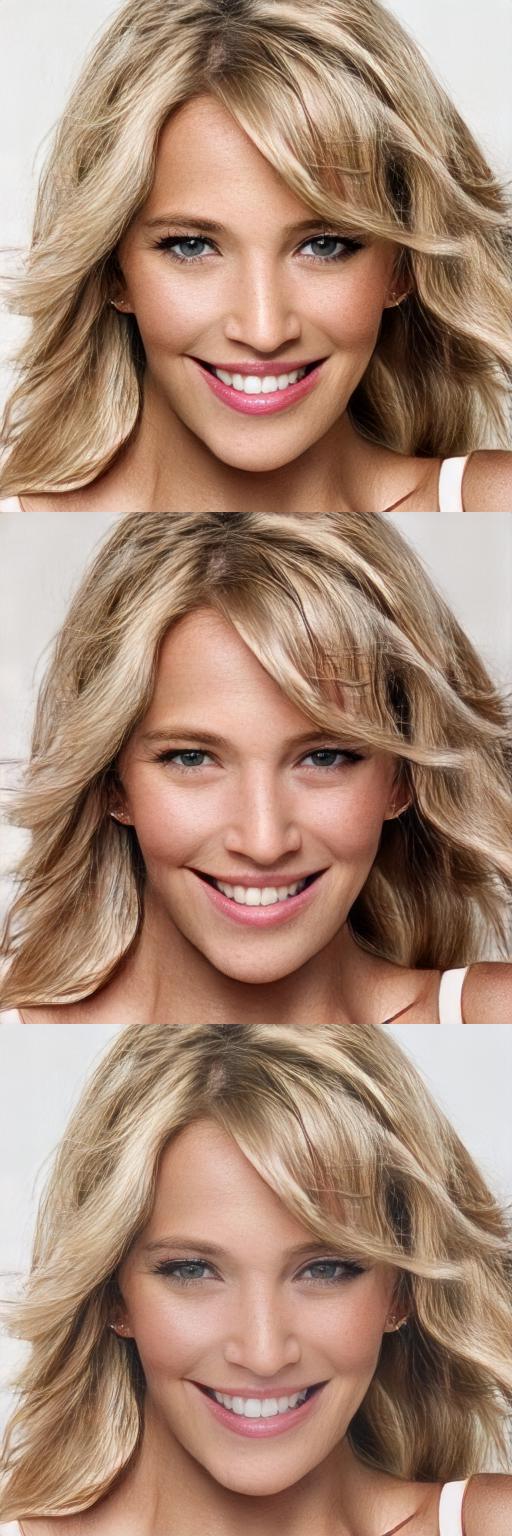}
&
\includegraphics[width=\linewidth]{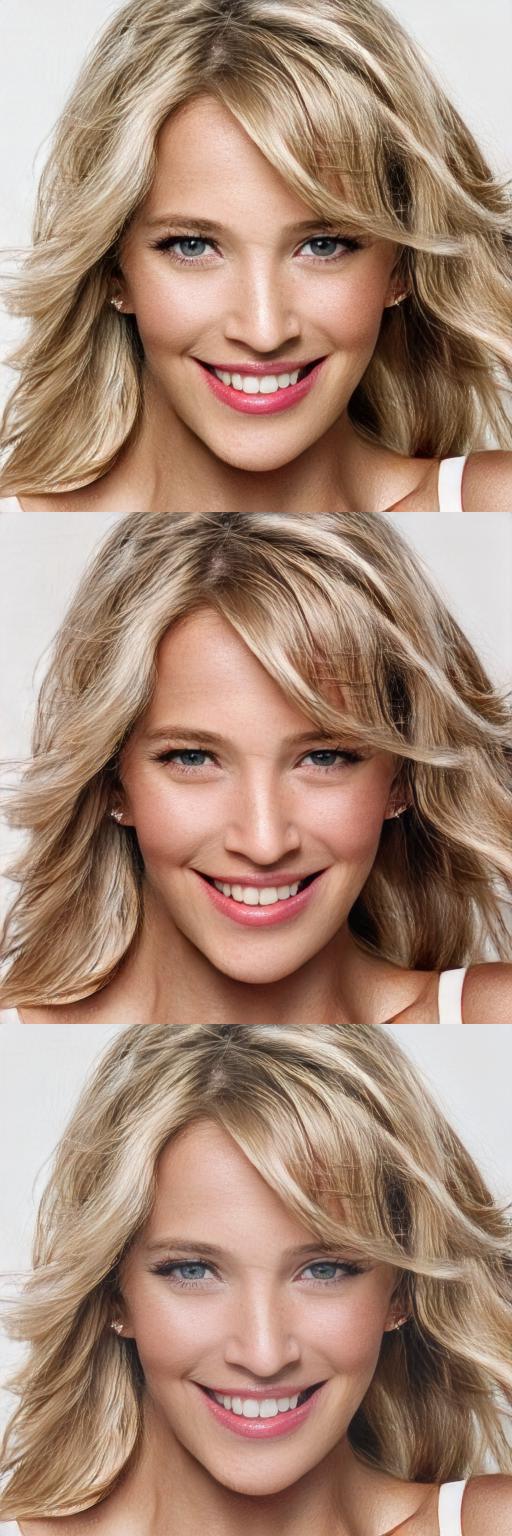}
\\
\end{tabular} 
\caption{\textbf{Choice of feature code insertion layer $K$.} The first column shows the source image (yellow frame) and two reference images for style mixing.
In the second to last column, the first row is the inversion results of each configuration, the second and third rows are the style mixing results, generated from the feature code of the source image and the latent code of the reference image.
Choosing $K=4$ yields good style mixing effects but lower reconstruction quality. Choosing $K=6$ or $7$ encodes nearly all the information in the feature code, which is limiting for editing. Our choice of $K=5$ holds a balanced trade-off between editing capacity and reconstruction quality.
}
\label{choice_k}
\end{figure*}

\noindent \textbf{Choice of $K$} \quad 
We provide an additional ablation study on the choice of feature code insertion layer $K$.
We compared $K=4$, $K=6$ and $K=7$ with our baseline $K=5$.
A different model is trained for each configuration. Figure \ref{choice_k} shows the qualitative results of inversion and style mixing. We observe that using $K=4$ yields good style mixing effects but lower reconstruction quality. While choosing $K=6$ or $7$ generates perfection reconstruction, the style mixing effects is less obvious. Using $K=6$ or $7$ encodes nearly all the information in the feature code, thus limited in editing. Our choice of $K=5$ holds a balanced trade-off between editing capacity and reconstruction quality.

\section{Limitations}

\begin{figure*}[t]
\centering
\small
\setlength{\tabcolsep}{0pt}
\renewcommand{\arraystretch}{0}
\begin{tabular}{P{0.124\linewidth}P{0.124\linewidth}P{0.124\linewidth}P{0.124\linewidth}P{0.124\linewidth}P{0.124\linewidth}P{0.124\linewidth}P{0.124\linewidth}}
\centering
Source & Inversion & Age & Gender & Smile & Eyeglasses & Chubby & Pose
\\
\multicolumn{8}{c}{\includegraphics[width=0.99\linewidth]{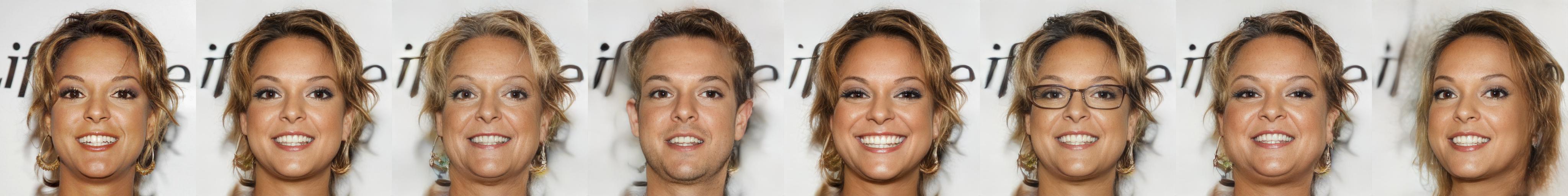}}
\\
\multicolumn{8}{c}{\includegraphics[width=0.99\linewidth]{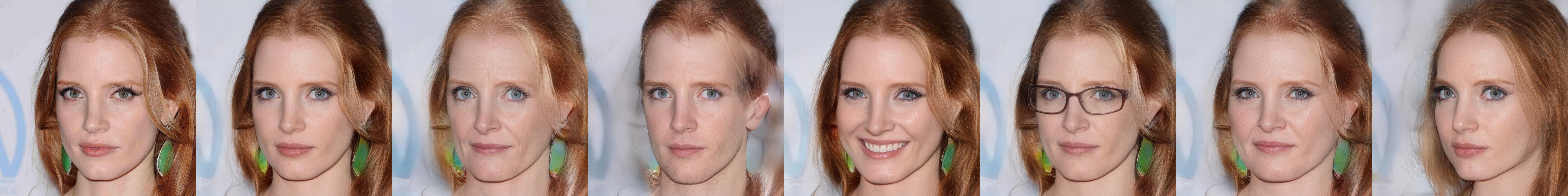}}
\\
\multicolumn{8}{c}{\includegraphics[width=0.99\linewidth]{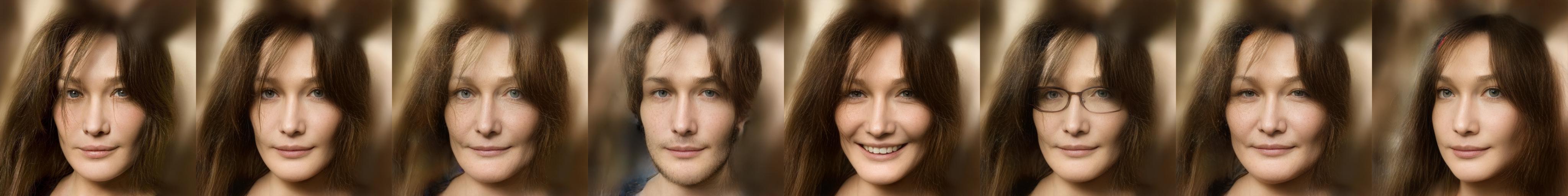}}
\\
\multicolumn{8}{c}{\includegraphics[width=0.99\linewidth]{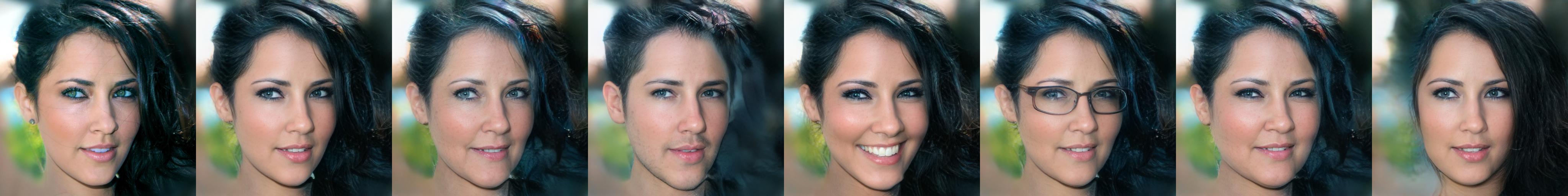}}
\\
\multicolumn{8}{c}{\includegraphics[width=0.99\linewidth]{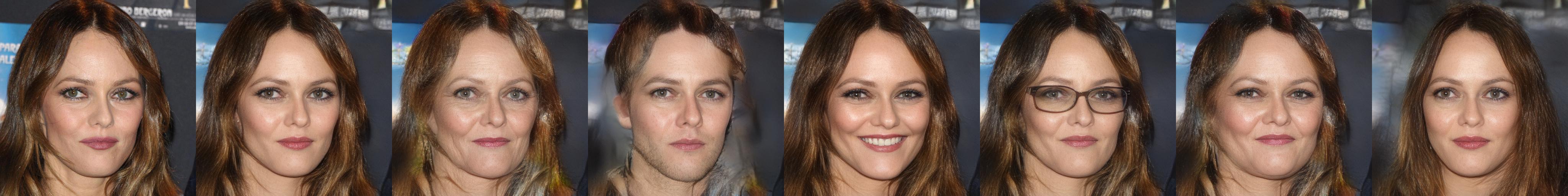}}
\\
\multicolumn{8}{c}{\includegraphics[width=0.99\linewidth]{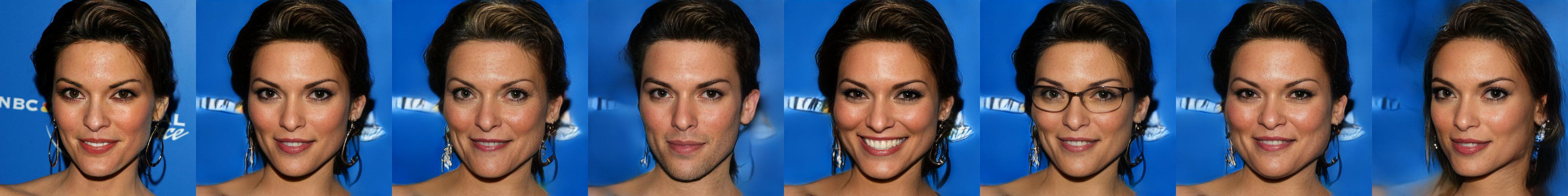}}
\\
\multicolumn{8}{c}{\includegraphics[width=0.99\linewidth]{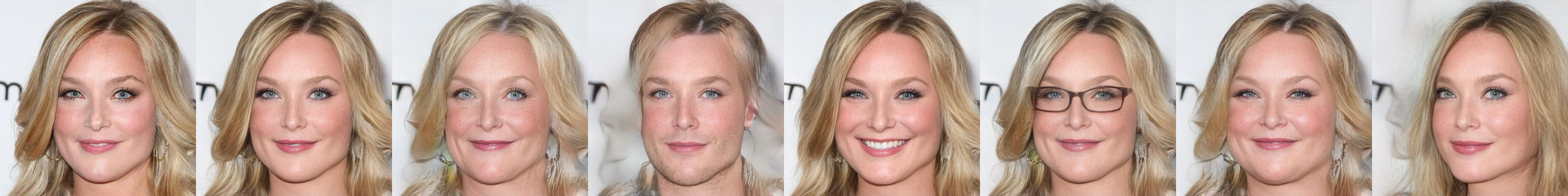}}
\\
\multicolumn{8}{c}{\includegraphics[width=0.99\linewidth]{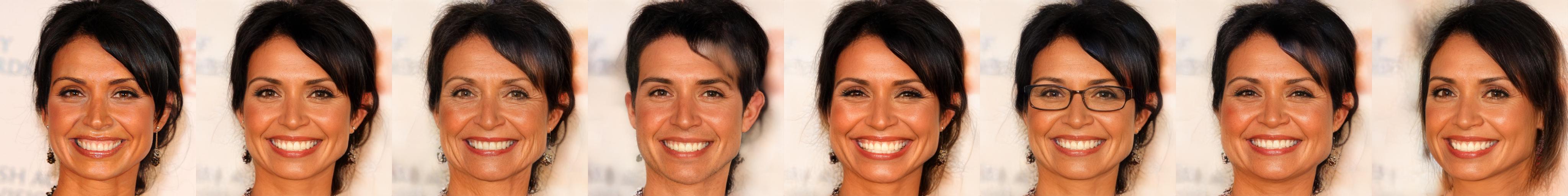}}
\\
\end{tabular}
\caption{\textbf{Editing on face domain}. 
We show additional facial attribute editing results. The latent editing directions are computed using InterFaceGAN~\cite{shen2019interpreting}, except the last attribute `pose', computed with SeFa~\cite{shen2021closed}.
}
\label{attr_edit}
\end{figure*}

In this section, we discuss the limitations of our proposed method.
Our encoder learns to project an image to a feature code and a latent code. To perform latent space editing, it is necessary to modify the feature code using equation~$(1)$ in the main paper. We have noticed that in some cases, this may be not sufficient. For instance in Figure \ref{attr_edit}, the editing results of attribute `gender' is less satisfying when the original person has long hairs. In the future, it could be helpful to study further improvements for the feature code editing, \eg, by including masks for interested area, or modifying only the relevant channels to achieve local editing~\cite{collins2020editing}.



\end{document}